%% file: draft.tex
\documentclass{article}
\usepackage[numbers]{natbib}
\usepackage{amsmath,amssymb}
\usepackage{hyperref}
\usepackage{graphicx}
\usepackage{colortbl}
\usepackage[section]{algorithm}
\usepackage{algpseudocode}
\algtext*{EndWhile}
\algtext*{EndIf}
\algtext*{EndFor}
\usepackage[normalem]{ulem}

\definecolor{Red}{rgb}{0.94, 0.12, 0.14}
\definecolor{Green}{rgb}{0.007, 0.62, 0.53}
\definecolor{Yellow}{rgb}{0.98,0.98,0.42}
\usepackage{typearea}
\typearea{12}

\begin{document}
\let\WriteBookmarks\relax
\def\floatpagepagefraction{1}
\def\textpagefraction{.001}

\title{Is Selection All You Need in Differential Evolution?}  

\author{Tomofumi Kitamura and Alex Fukunaga}

\maketitle

\section{Introduction}
Evolutionary computation is a powerful method for black-box optimization problems.
Evolutionary computation has been applied to numerous domains, including optimization of weather radar networks~\cite{kurdzo2012objective}, radar system design~\cite{egger2017radar}, and precipitation nowcasting when combined with machine learning techniques~\cite{csen2001genetic, ngan2023hybrid, lee2006parameter, kim2024development}, and other many industries~\cite{kumar2020test}.

In {\it single-objective optimization}, multiple continuous-valued variables are optimized to minimize a single objective function. A single-objective optimization problem aims to find a real-valued vector $x = (x_1, \cdots, x_D)$, where $D$ is the dimensionality of the problem, that minimizes an objective function $f: \mathbb{R}^D \to \mathbb{R}$. The value $f(x)$ is referred to as the fitness of solution $x$. The global optimum $x^*$ is defined as satisfying $f(x^*) \leq f(x), \forall x \in \mathbb{R}^D$.

When evaluating the objective function $f(x)$ involves complex simulations \cite{oyama2017simultaneous, sato2008low, shimoyama2011multi} or when $f$ is not analytically computable \cite{sims1991artificial}, algorithms that do not require access to the internal structure of $f$ are necessary. Such scenarios, where only the output of the objective function is available, fall under the category of {\it black-box single-objective optimization}.

Differential Evolution (DE) is a widely used algorithm for black-box optimization problems. DE generates new candidate solutions $u_i^t$ at each generation $t$—a discrete time step in the evolutionary process—through operations known as mutation and crossover. The fitness of each new candidate is evaluated, and if $u_i^t$ outperforms the existing candidate $x_i^t$, $u_i^t$ replaces $x_i^t$ in the next generation as $x_i^{t+1}$. If not, the existing candidate survives, and $u_i^t$ is discarded. This process of {\it generational replacement} based on fitness is a fundamental concept common to various evolutionary algorithms such as Genetic Algorithms (GA), and has been a standard feature of DE since the original DE formulation \cite{StornP96}.

However, in modern DE implementations, a major challenge lies in the limited population diversity caused by the fixed population size enforced by the generational replacement.
Population size is a critical control parameter that significantly affects DE performance. Larger populations inherently contain a more diverse set of individuals, thereby facilitating broader exploration of the search space. For high-dimensional problems, such as those with 50 dimensions, larger population sizes are often employed compared to lower-dimensional problems \cite{TanabeF14CEC}. Conversely, when the maximum evaluation budgets is constrained, smaller populations focusing on a limited number of promising candidates may be more suitable.
Many state-of-the-art DE variants incorporate an archive mechanism \cite{TanabeF14CEC, stanovov2022nl}, in which a subset of discarded individuals is preserved in an archive $A^t$ during generation replacement and reused in mutation operations. This practice increases diversity by expanding the candidate pool to $P \cup A$. However, maintaining what is essentially a secondary population via an archive introduces additional design considerations, such as policies for insertion, deletion, and appropriate sizing.

We observe that much of the increasing complexity of state-of-the-art DE can be ascribed to three, widely accepted assumptions/tenets underlying DE implementation:
(1) individual are replaced by offspring (either by direct offspring or by the offspring of other individuals);
(2) population sizes are either fixed, or are reduced as search progresses (to focus search); and
(3) ``failed'' individuals (offspring with worse fitness than their parents) are discarded.

Underlying all three of these assumptions is the central assumption that {\it DE should throw away much of the information discovered during search} (i.e., the individuals created and evaluated during search)
-- the replacement policy eliminates parents, population size reduction eliminates members from the population, and discarding failed individuals means that such failed individuals will not further contribute to search progress.
In fact, by the end of the search, all that remains in a standard DE is the individuals in the final population, which is a very small fraction of the individuals that have been created and evaluated, and almost all the ``knowledge'' about the search space has been thrown away.

In this paper, we question these assumptions, and explore a fundamentally different design for DE which starts with the premise: {\it What if we did not discard any information (individuals) ?}
We propose a novel DE framework called Unbounded Differential Evolution (UDE), which adds all generated candidates to the population without discarding any based on fitness. Unlike conventional DE, which removes inferior individuals during generational replacement, UDE eliminates the need for replacement altogether, along with the associated complexities of archive management and dynamic population sizing. UDE represents a fundamentally new approach to DE, relying solely on selection mechanisms and enabling a more straightforward yet powerful evolutionary process.

We use UDE as a framework for
``deconstructing'' state-of-the-art DE and reconsider whether the complexity of modern DE variants is necessary.
Instead ofgenerational replacement, supplemental populations (archives), and deterministic population size reduction strategies, perhaps many of the standard components of a DE are actually unnecessary and can be subsumed by the process of selection. Instead of carefully designing mechanisms for discarding information, maybe we confocus on designing selection operators which effectively choose from all of the information (individuals) generated during the search -- i.e.,  {\it Is selection all you need?}

The rest of the paper is organized as follows.
Section \ref{sec:background} reviews preliminaries and related work on DE. 
In Section \ref{sec:udeall}, we propose Unbounded DE (UDE), as well as variants of UDE which incorporate parameter adaptation.
We experimentally evaluate UDE in Section \ref{sec4}.
We show that UDE and its adaptive variant, USHADE, is competitive with standard adaptive DE such as SHADE and LSHADE.
We explore the simulation of population size adaptation in the UDE framework, and  show that selection policies can be used to mimic population size increases as well as decreases (Section \ref{simulating-popsize}). We also explore the necessity of discarding ``failed'' individuals with worse fitness than their parents, and show that the standard DE practice of discarding failed individuals is unnecessary in the UDE framework  (Section \ref{sec:ex2}).
Section \ref{sec:last} concludes with a discussion and directions for future work.

This work significantly extends and expands upon work preliminary results presented in a CEC2022 paper \cite{kitamura2022differential}.
The earlier paper proposed and focused on a variant of UDE which discarded failed individuals which scored worse than their parents (the variant referred to as USHADE/DF in Sections \ref{sec:ude-without-failed-individuals} and \ref{sec:ex2} of this paper). The main versions of UDE and USHADE proposed in this paper keeps all individuals in the population, and  are new in this paper.
Almost all text in the paper is completely newly written -- the presentation of UDE and its variants has been completely rewritten, as well as the survey of previous work.
Most of the experiments are completely new, specifcally: evaluation on the CEC2022 benchmarks (Section \ref{sec:cec22}), analysis of robustness of search with respect to maximum evaluation budgets (Section \ref{sec:robustness-evaluation-budgets}), and 
analysis of the role and usage of failed individuals (Section \ref{sec:ex2}) are completely new to this paper.
All experimental results, figures, and tables in this paper are new, and do not overlap with results in \cite{kitamura2022differential}.

\section{Preliminaries and Background} \label{sec:background}
\input{background.tex}

\input{recent_de.tex}
\section{Proposed method} \label{sec:udeall}
\input{ude-jp.tex}

\section{Experimental Evaluation}  \label{sec4}
\input appendix-1

\subsection{Performance Evaluation: Is Replacement Necessary?} \label{sec:cec22}

We evaluate the following algorithms, all of which use current-to-pbest mutation, and binomial crossover.
All algorithms were evaluated on the CEC2014 benchmarks, 51 independent runs per problem, and the maximum evaluation budgets on each run was $20,000 \times D$.

\begin{itemize}
\item Baseline-DE (Alg.~\ref{alg:DE}): A baseline DE based on the classical DE of \cite{StornP96}. This uses fixed parameter values for scale factor ($F=0.5$), crossover rate ($C=0.5$) and a population size ($|P^t|=100$).
Pbest rate $p=0.11$. There is no archive. %
  
\item UDE(DPT) (Alg.~\ref{alg:uDE}): UDE {\it without} parameter adaptation, using the DPT tournament policy (Section \ref{sec:ex1}) with tournament parameter ($T=round(|P^t|/|P^1|)$), initial population size $|P^1|=18\times D$.
This uses fixed parameter values for scale factor ($F=0.5$), crossover rate ($C=0.5$), 
and generates a fixed number ($gensize=100$) of individuals each generation.
A population size $|P^t|$ grows monotonicaly, pbest rate $p=0.11$. %

\item SHADE (Alg.~\ref{alg:SHADE},~\cite{TanabeF13}): Adaptive DE using success-history based parameter adaptaion of $F$ and $C$.
As in the implementation~\cite{lshadecode} by the original author \cite{TanabeF13}, the success-history parameters $M_F$ and $M_C$ are vectors of length $H=D$, pbest rate $p=0.10$, population size $|P^t|=100$, and archive size $|A_\text{max}|=2\times|P|$. %
  
\item LSHADE (Sec.~\ref{sec:LPSR},~\cite{TanabeF14CEC}: SHADE with Linear Population Size Reduction.
Following the implementation~\cite{lshadecode} by the original author \cite{TanabeF14CEC}, 
  initial population size $|P^1|=18 \cdot D$, population size is reduced  using eq.~\ref{eq:LPSR}, history length $H=6$, and archive size $|A^t_\text{max}|=1.4\cdot|P^t|$.

\item USHADE(T) (Alg.~\ref{alg:uADE}): UDE with success-history based parameter adaptation using the T tournament selection operator.
This generates a fixed number ($gensize=100$) of individuals each generation,
and population size $|P^t|$ grows monotonicaly.
Other parameters ($|P^1|$, $H$, $p$) are the same as LSHADE.
  
\item USHADE(DPT) (Alg.~\ref{alg:uADE}): UDE with success-history based parameter adaptation using the DPT tournament selection operator (Section \ref{sec:ex1}). All parameters are the same as USHADE(T).

\end{itemize}

\subsubsection{Evaluation of UDE without parameter adaptation: Baseline DE vs. UDE(DPT) }

First, we evalute whether the key, novel elements of our approach (unbounded population, no generational replacement) are viable compared to standard DE.
Thus, we compare the performance of UDE(DPT) against baseline-DE, neither of which incorporate parameter control.

\input tab-DE

Table~\ref{tab:DE} shows the Wilcoxon ranked sum test (p=0.05) comparison of the fitnesses values reached by Baseline-DE vs. UDE(DPT). Overall, UDE(DPT) outperformed baseline-DE across a wide range of functions at all dimensions $D=10,30,50$. $F14$ is the only function where baseline-DE consistently outperforms UDE(DPT).
Comparing the ECDFs of Baseline-DE and UDE(DPT) in Figure~\ref{fig:cec14}, the UDE ECDF curve is clearly above the DE curve for $D=10,30,50$.

Thus, our results clearly indicate that UDE(DPT) is competitive with Baseline-DE -- given the same mutation and crossover operators, the unbounded population approach of UDE outperforms generational replacement approach of standard DE.

\subsubsection{Evaluation of UDE with parameter adaptation}

\begin{figure}[htbp]
\begin{center}
\includegraphics[width=.92\textwidth]{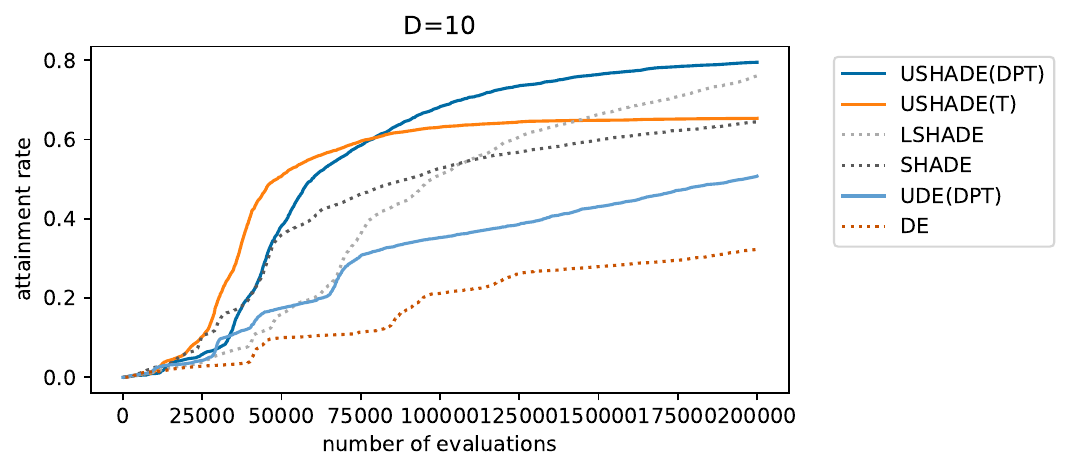}
\includegraphics[width=.92\textwidth]{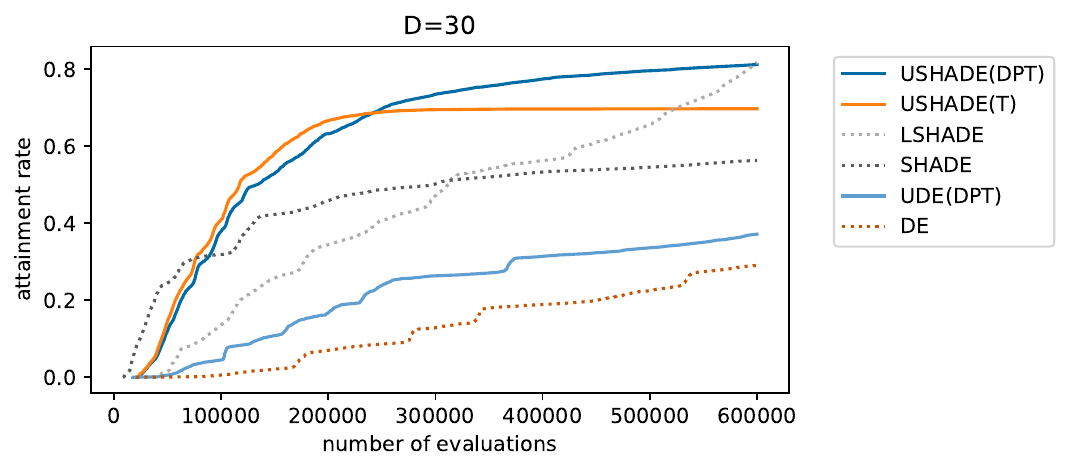}
\includegraphics[width=.92\textwidth]{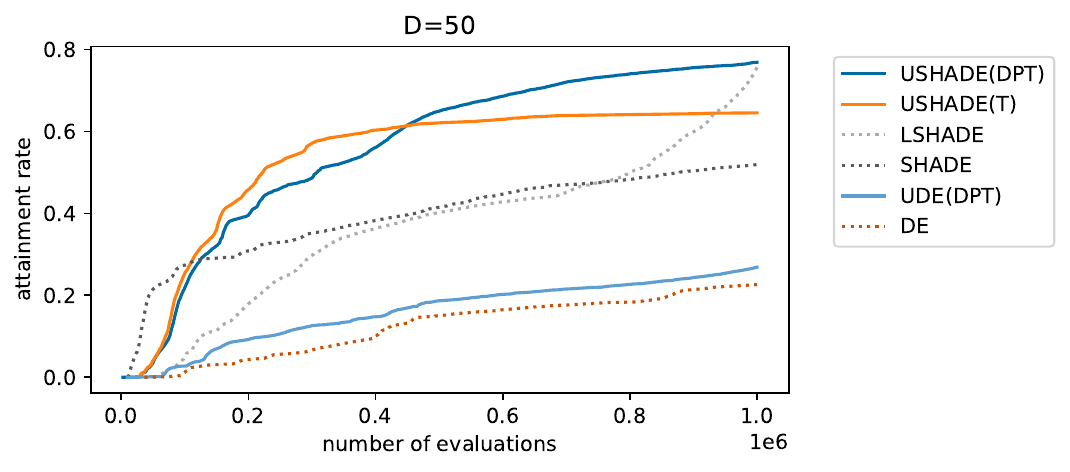}
\end{center}
\caption{\small
  ECDFs of six algorithms %
  on the CEC 2014 benchmark suite for $D=10, 30, 50$. The x-axis indicates the maximum evaluation budgets, and the y-axis shows the proportion of 51 trials that attained the target value at each point. For each problem, three targets were defined based on the median and first/third quartiles of $\FinalBestSoFar$, calculated across all six algorithms. The ECDFs represent the average attainment rate over 30 benchmark functions.}
\label{fig:cec14}
\end{figure}

We now compare UDE with adaptation (USHADE) with the standard adaptive DEs SHADE and LSHADE.
Figure~\ref{fig:cec14} shows ECDFs of the algorithms' average attainment rates. USHADE(DPT) outperformed LSHADE for all $D$.
In the early stage of search (approximately up to 10\% of the maximum evaluation budgets), SHADE tended to be the most effective algorithm,  %
but by the midpoint of the search, is overtaken by USHADE(T), USHADE(DPT), and LSHADE.
Overall, after the midpoint of the search, USHADE(DPT) has the highest ECDF curve among the algorithms.

Tables \ref{tab:uadeT-D10}-\ref{tab:uadeT-D50} compare $\FinalBestSoFar$ (i.e., after $2e4\times D$ evaluations) using the Wilcoxon rank-sum test ($p=0.05$).
USHADE(DPT) outperformed SHADE was worse on 16, 19, and 22 problems for $D=10,30,50$, respectively. USHADE(DPT) outperformed LSHADE on 10, 11, and 9 problems, while LSHADE outperformed USHADE(DPT) on 6, 8, and 11 problems for $D=10,30,
50$, respectively.

For multimodal functions $F10$ to $F15$, LSHADE generally performed better than USHADE(DPT). For hybrid and composite functions ($F17$ onward), USHADE(DPT) outperformed LSHADE at $D=10$, while at $D=30$ and $50$, the two algorithms performed comparably  with an approximately equal number of wins and losses.

\input tab-alladd-D10
\input tab-alladd-D30

Comparing the USHADE tournament policies T and DPT,
the USHADE(DPT) ECDF curves are consistently above the USHADE(T) curves.
At the end of search, USHADE(DPT) outperformed USHADE(T) on 16, 12 and 14 problems for $D=10, 30, 50$, respectively, and USHADE(DPT) was outperformed by USHADE(T) on 3, 2 and 2 problems for $D=10, 30, 50$ respectively.

Thus, the results on the CEC2014 benchmarks show that:
(1) USHADE(DPT) is competitive with SHADE and LSHADE, and
(2) the DPT tournament selection strategy significantly outperforms the baseline T strategy, indicating the usefulness of designing tournament policies to maintain diversity.

\clearpage
\input tab-alladd-D50

\subsubsection{Evaluation on CEC2022 benchmarks}
\label{sec:cec2022}

Finally, we compared the performance of USHADE(DPT) with state-of-the-art methods on the CEC2022 benchmarks.
Specifically, we compare USHADE(DPT) with the top four algorithms in CEC 2022 competition:

\begin{itemize}
\item EA4eig~\cite{bujok2022eigen} (1st place), an
ensemble of four high-performing evolutionary algorithms:
(1) (Covariance Matrix Adaptation Evolutionary Strategy (CMA-ES), (2) Differential Evolution with Covariance Matrix Learning and Bimodal Distribution Parameter Setting (CoBiDE), (3) an adaptive variant of jSO~\cite{brest2017single}, and (4) Differential Evolution With an Individual-Dependent Mechanism (IDE)) enhanced by Eigen crossover.
  
\item NL-SHADE-LBC~\cite{stanovov2022nl} (2nd place), a LSHADE variant which incorporates finely tuned biased parameter adaptation and combines the rank-based mutation strategy introduced in LSHADE-RSP~\cite{stanovov2018lshadersp}.

\item NL-SHADE-RSP-MID~\cite{biedrzycki2022version} (3rd place), a LSHADE variant with a restart mechanism. Generates individuals biased to the center of subpopulations obtained by partitioning the population by k-means clustering.
\item S-LSHADE-DP~\cite{van2022dynamic} (4th place), a LSHADE variant which monitors its population using indicators of stagnation, and when stagnation is detected, it attempts to resolve the stagnation by generating new individuals using perturbations different from mutation.
\end{itemize}

EA4eig is an ensemble algorithm, while the other three are variants of LSHADE.

We compared the distributions of the final best-so-far values achieved by the algorithms on the CEC2022 benchmark problems.
For USHADE(DPT), following the CEC 2022 competition protocol, we performed 30 independent runs on 12 functions, with maximum evaluation budgets of 200,000 and 1,000,000 for $D=10,20$, respectively. 
For EA4eig, NL-SHADE-LBC, NL-SHADE-RSP-MID, and S-LSHADE-DP, we use the raw data for the final best-so-far values downloaded from the CEC2022 competition data repository \cite{cec2022}.

\input tab-cec22comp

Table \ref{tab:cec22} shows the results of the comparisons.
For $D=10$, USHADE(DPT) outperformed the third-place algorithm from the CEC 2022 competition on 3 problems and tied (no statistically significant difference) on  4 others. Compared to the fourth-place algorithm, USHADE(DPT) performed better on 2 problems and tied on 6 problems. The first- and second-place algorithms achieved the same optimal solutions as USHADE(DPT) and outperformed USHADE(DPT) on 7 and 6 problems, respectively.
For $D=20$, USHADE(DPT) outperformed the third-place algorithm on 5 problems and was worse on 4. Against the fourth-place algorithm, USHADE(DPT) was superior on 3 problems and worse on 3. In comparison with the first- and second-place algorithms, USHADE(DPT) performed better  on 1 and 2 problems, respectively, reached the same optimal solution on 3 problems, and was outperformed on 5.
Thus, USHADE(DPT) performed worse than the 1st and 2nd place entries (EA4eig, NL-SHADE-LBC) in the CEC2022 competition, but comparably to the 3rd and 4th place entries (NL-SHADE-RSP-MID, S-LSHADE-DP).

\input{sec5.tex}

\section{Conclusion} \label{sec:last}

This paper proposed Unbounded DE, a novel approach to DE which never discards or replaces individuals, and adds all created individuals to an unbounded population.
We sought to develop a simplified search framework where a single selection operator subsumes
the role of standard components of DE -- generational replacement, auxiliary populations (e.g., archives), and population resizing.

The main contribution of the paper are as follows:

\begin{itemize}
\item 
  We showed that a simple implementation UDE is competitive with DE, and that UDE which incorporates the parameter adaptation scheme used by SHADE is competitive with SHADE and LSHADE. In other words, we show that a fixed population size and generational replacement is not a necessary feature of DE, and can be substituted by an unbounded population and selection mechanisms.
\item We showed that in UDE, adaptive population resizing can be simulated using selection policies. Unlike standard DE population resizing techniques such as linear reduction \cite{TanabeF14CEC}, UDE can effectively simulate both ``increase'' the population as well as ``decreasing'' the population, and also  does not require a predetermined schedule. This simplicity and adaptivity is an advantage over state-of-the art DE approaches that rely on explicit stagnation detection  or pre-scheduled deterministic population reduction schedules (c.f., \cite{van2022dynamic,stanovov2022nl,biedrzycki2022version}).

\item While standard DE variants discard ``failed'' individuals which have worse fitness than their parents, we showed that in UDE, discarding ``failed'' individuals is unnecessary, and that the failed individuals can sometimes enhance performance.

\end{itemize}

The comparison of USHADE with state-of-the-art DEs from the CEC2022 benchmark competition (Section \ref{sec:cec2022}) shows that UDE is a promising direction.
While the state-of-the-art solvers participating in the CEC2022 competition were (1) tuned for the CEC2022 competition protocol (e.g., population reduction schedules were tuned for the maximum evaluation budgets) and (2) used recently developed various methods (adaptation mechanisms, mutation operators),
In contrast, USHADE(DPT) has not been tuned.
The only control parameter unique to USHADE, $gensize$, was set to 100 (same simple default value as for the fixed population size for SHADE in the original 2013 paper \cite{TanabeF13}.
All other control parameters (initial population size $|P^1|$, history length $H$, pbest rate $p$) were the default settings for LSHADE from the original 2014 paper \cite{TanabeF14CEC}.
Furthermore, USHADE relies on the same classical mutation and adaptation mechanisms from 2013 as SHADE.

Nevertheless, USHADE(DPT) performed comparably to the 3rd/4th place algorithms, showing that 
that the UDE approach (unbounded population, no generational replacement) seems to be a viable approach. 
Improving UDE by incorporating more modern parameter adaptation methods and mutation strategies, as well as parameter tuning ($gensize$, $|P^1|$) are direction for future work.

Improving the selection policy is another direction for future work.
Considering individual success as described in Sec.~\ref{sec:no-use} may improve search performance in higher dimensions.
In addition, combining multiple selection policies in the selection of individuals or ensembles of selection policies are another interesting direction.

\end{document}

%% file: background.tex
\subsection{Differential Evolution (DE )}
Differential Evolution (DE) \cite{StornP96, storn1997differential} is a population-based optimization algorithm that iteratively improves a population $P^t$ of candidate solutions, where each individual $x$ is a $D$-dimensional vector. The index $t$ denotes the generation number, and the population size is denoted by $|P^t|$.
Algorithm~\ref{alg:DE} shows the pseudocode for the canonical DE algorithm. In line 2, the initial population $P^1$ is generated, typically by sampling each individual uniformly at random from the search space. The fitness of all individuals $f(x^{i,1})$ is evaluated in lines 3 and 4. In each subsequent generation, $|P^t|$ offspring $u^{i,t}\: (i=1,...,|P^t|)$ are generated and evaluated.
Mutation is performed in line~\ref{alg:DE-mutation}, where a mutant vector $v^{i,t}$ is generated using the {\it difference} of two individuals from the population $P^t$ -- this is a defining feature that gives DE its name. Crossover (line~\ref{alg:DE-crossover}) then combines the parent $x^{i,t}$ and the mutant $v^{i,t}$ to form the offspring $u^{i,t}$. The offspring is evaluated in line 10.
Lines 11 to 14 implement selection. If $f(u^{i,t}) \leq f(x^{i,t})$, the offspring replaces the parent in the next generation. This process of mutation, crossover, evaluation, and selection repeats until a termination condition is met.
Various mutation and crossover strategies exist for DE, as discussed in the following sections. In contrast, the selection mechanism (lines 11–14) is common to all DE variants.

The control parameters for DE are summarized in Table~\ref{tab:3parameters}. As observed in prior studies~\cite{gamperle2002parameter, mezura2006comparative}, the performance of evolutionary algorithms is generally sensitive to these parameters. Section~\ref{sec:adaptive-DE} discusses adaptive DE algorithms that dynamically adjust the scale factor $F$ and crossover rate $C$ during the optimization process. Section~\ref{sec:LPSR} addresses strategies for decreasing the population size $|P^t|$ over time.
In adaptive DE, $F$ and $C$ are typically constrained to $(0,1]$ and $[0,1]$, respectively. When fixed values are used for the purpose of evaluating algorithm performance independently of parameter adaptation, we follow conventional, setting $F=0.5, C=0.5$.
  The parameter $p$, known as the pbest rate, is used in the current-to-pbest mutation strategy~\cite{ZhangS09} (see below) and is generally fixed.
  The archive size $|A|$ is relevant for DE variants employing an archive (see below).

\begin{table}[htbp]
\caption{\small
Control parameters for DE. $|P|, F, C$ are the 3 control parameters for standard DE. The archive size $|A|$ is used by DE varaints which use archives, and $p$ is the pbest rate for current-to-$pbest$ mutation.}
\label{tab:3parameters}
\centering
\begin{tabular}{lll}\hline
Control Parameter & Symbol & Description \\\hline\hline
Population size       & $|P^t|$ & integer $\ge4$ because standard DE mutation involve 4 individuals\\
Scale factor   & $F$ & Real number in (0,1]\\
Crossover rate & $C$ & Real number in [0,1]\\
pbest rate & $p$ & Used in current-to-pbest mutation \cite{ZhangS09}, real number in [0,1]\\
Archive size & $|A^t|$ & Used in current-to-pbest mutation \cite{ZhangS09}, integer $\geq 0$\\
\hline
\end{tabular}
\end{table}

\begin{algorithm}[htbp]
\caption{Standard DE \cite{StornP96}}
\label{alg:DE}
\begin{algorithmic}[1]
\State $t = 1$;
\State Initialize population $P^t$; 
\For{ $i = 1$ to $|P^t|$}
    \State evaluate $f(x^{i,t})$;
\EndFor
\While{not termination condition}
\For{ $i = 1$ to $|P^t|$}
    \State generate $v^{i,t}$ using mutation; \label{alg:DE-mutation}
    \State generate $u^{i,t}$ using crossover between $v^{i,t}$ and $x^{i,t}$; \label{alg:DE-crossover}
\EndFor
\For{ $i = 1$ to $|P^t|$}
    \State evaluate $f(u^{i,t})$;
    \If{$f(u^{i,t}) \leq f(x^{i,t})$}
        \State $x^{i,t+1} = u^{i,t}$;
    \Else
        \State $x^{i,t+1} = x^{i,t}$;
    \EndIf
\EndFor
\State $t = t+1$;
\EndWhile
\end{algorithmic}
\end{algorithm}

\paragraph{Mutation (Alg.~\ref{alg:DE}, line~\ref{alg:DE-mutation})} \label{sec:mutation}

A basic mutation strategy used in the original DE \cite{StornP96} is rand/1 \cite{StornP96}, defined as:
\begin{eqnarray}
v^{i,t} = x^{r1,t} + F \cdot (x^{r2,t} - x^{r3,t}) \label{eq:rand1}
\end{eqnarray}
where $x^{r1,t}, x^{r2,t}, x^{r3,t}\: (r1\neq r2, r1\neq r3, r2\neq r3)$ are three distinct individuals randomly selected from the population $P^t$. The magnitude of $F$ scales the differential vector and thus influences the distance of the mutant $v^{i,t}$ from the base vector $x^{r1,t}$.
In rand/1, the standard approach to selecting invididuals from the population  is uniform random sampling.

The  current-to-pbest mutation strategy, introduced in JADE \cite{ZhangS09} is widely used in state-of-the-art DE algorithms. It is defined as:
\begin{eqnarray}
v^{i,t} = x^{i,t} + F \cdot (x^{\text{pbest},t} - x^{i,t}) + F \cdot (x^{r1,t} - x^{r2,t}) \label{eq:cur-to-pbest}
\end{eqnarray}
Here, $x^{\text{pbest},t}$ is selected uniformly at random from the top $|P^t| \cdot p$ individuals in the population ranked by fitness. Compared to rand/1, this strategy biases the search toward promising regions of the solution space.

\paragraph{Crossover (Alg.~\ref{alg:DE}, line~\ref{alg:DE-crossover})} \label{sec:crossover}
The offspring $u^{i,t}$ is generated through crossover (Algorithm~\ref{alg:DE}, line~\ref{alg:DE-crossover}), which recombines the parent $x^{i,t}$ and the mutant $v^{i,t}$ on a per-dimension basis. Binomial crossover \cite{storn1997differential}, a widely used scheme, is  defined as:
\begin{eqnarray}
u^{i,t}_{j}=\left\{ \begin{array}{ll}
v^{i,t}_{j}  & \text{if} \; u \le C \; \text{or} \; j= j_\text{rand} \quad \text{where} \; u \sim U(0,1)\\
x^{i,t}_{j}  & \text{otherwise} \label{eq:crossover}
\end{array} \right.
\end{eqnarray}
A uniform random number $u$ determines whether each dimension $j$ inherits from the mutant or the parent. The dimension $j_\text{rand}$ is always inherited from the mutant to ensure that $u^{i,t}_{j_\text{rand}} = v^{i,t}_{j_\text{rand}}$.

\subsection{Adaptive DE} \label{sec:adaptive-DE}
The performance of DE, like other evolutionary algorithms, is highly influenced by control parameter settings~\cite{gamperle2002parameter, mezura2006comparative}. Classical DE algorithms use fixed values for key parameters such as the scale factor $F$ and crossover rate $C$. However, optimal values for these parameters vary depending on the problem characteristics and search dynamics. Adaptive parameter control can significantly improve performance.
For example, larger values of $F$ facilitate exploration by amplifying differential vectors, which is beneficial in the early stages of multimodal optimization. Conversely, reducing the value of $F$ focuses the search. When $C$ is close to 0, offspring tend to inherit values from their parents in many dimensions, promoting optimization on a per-dimension basis, so  in problems without inter-variable dependencies, setting a small $C$ allows the algorithm to exploit this independence during the search. On the other hand, increasing $C$ enables offspring to have values not present in the parents, which tends to enhance performance on problems involving interdependent variables.

Parameter control in evolutionary computation is typically classified into three categories: deterministic, adaptive, and self-adaptive control~\cite{eiben1999parameter}.

Adaptive control, dynamically adjusts parameters based on feedback from the search process, allowing parameter values to align more closely with the characteristics of the problem at hand. For $F$ and $C$, adaptive control is commonly used. One widely adopted adaptive control method is success-history based control, described in Section~\ref{sec:parameter}.

In contrast, deterministic control modifies parameters according to predefined rules, independent of search progress. One example, linear population size reduction, is detailed in Section~\ref{sec:LPSR}, decreases population size $|P^t|$ linearly. Although an obvious disadvantage of deterministic control is lack of flexibility and inability to response to conditions during the search, deterministic control is widely used for population size control due to the difficulty in defining success or failure in population size.

Self-adaptive control evolves the parameters themselves as part of the individual genomes. However, self-adaptive control is rarely employed in state-of-the-art DE algorithms.

\subsubsection{Success-history based parameter adaptation (SHADE)} \label{sec:parameter}
Success-history-based parameter control has been widely adopted in state-of-the-art adaptive DEs since the introduction of SHADE~\cite{TanabeF13}, and is used in subsequent variants~\cite{TanabeF14CEC, stanovov2018lshadersp, gurrola2020colshade}. Algorithm~\ref{alg:SHADE} shows SHADE. The blue-highlighted lines are related to parameter control.
\begin{algorithm}[htbp]
\caption{SHADE (\textcolor{blue}{blue} is related to parameter control, and \textcolor{red}{red} is related to archives).}
\label{alg:SHADE}
\begin{algorithmic}[1]
\State $t = 1$;
\State Initialize population $P^t$; 
\textcolor{red}{\State $A^t = \emptyset$;} \label{algline:SHADE-archive}
\textcolor{blue}{\State Initialize contents of success histories $M_F$ and $M_C$ to 0.5; \label{algline:shade-init-histories}
\State $k = 1$; \label{algline:shade-init-k}
}
\For{ $i = 1$ to $|P^t|$}
    \State evaluate $f(x^{i,t})$;
\EndFor
\While{not termination condition}
\State \textcolor{blue}{$S_F = \emptyset$, $S_{C} = \emptyset$, $S_{\Delta f} = \emptyset$;}
\For{ $i = 1$ to $|P^t|$}
    \State \textcolor{blue}{select $r^i$ randomly from $\{1,\cdots,H\}$;} \label{algline:shade-choose-history-index}
    \While{\textcolor{blue}{$F^{i,t}\leq0$}}
        \State \textcolor{blue}{$F^{i,t}=\text{min}(\text{rand}_\text{cauchy}((M_F[r^i], 0.1),1)$;} \label{algline:shade-generate-f}
    \EndWhile
    \State \textcolor{blue}{$C^{i,t}=\text{max}(0,\text{min}(\text{rand}_\text{normal}(M_C[r^i], 0.1),1))$;} \label{algline:shade-generate-c}
    \State select $x^{pbest,t}, x^{r1,t}$ from $P^t$ and \textcolor{red}{select $x^{r2,t}$ from $P^t \cup A^t$}  $(i\neq r1, i\neq r2, r1\neq r2)$; \label{algline:SHADE-mutation}
    \State generate $v^{i,t}$ using current-to-{\it pbest} mutation (eq.~\ref{eq:cur-to-pbest});
    \State generate $u^{i,t}$ using binomial crossover (eq.~\ref{eq:crossover}) between $v^{i,t}$ and $x^{i,t}$;
\EndFor
\For{ $i = 1$ to $|P^t|$}
    \State evaluate $f(u^{i,t})$;
    \If{$f(u^{i,t}) \leq f(x^{i,t})$}
        \State $x^{i,t+1} = u^{i,t}$;
        \State \textcolor{blue}{$S_F = S_F \cup F^{i,t}$, $S_{C} = S_{C} \cup C^{i,t}$, $S_{\Delta f} = S_{\Delta f} \cup (f(x^{i,t}) - f(u^{i,t}))$;} \label{algline:shade-add-to-sets}
        \textcolor{red}{\If{$|A^t| < |A|_\text{max}$}
          \State $A^{t+1} = A^t \cup x^{i,t}$; \label{algline:SHADE-add1}
        \Else
          \State select $x^{r3,t}$ from $A^t$; \label{algline:SHADE-add2}
        \EndIf}
          \textcolor{red}{\State $A^{t+1} = (A^t \smallsetminus x^{r3,t}) \cup x^{i,t}$;} \label{algline:SHADE-add3}
    \Else
        \State $x^{i,t+1} = x^{i,t}$;
    \EndIf
\EndFor
\textcolor{blue}{\If{$S_F \neq \emptyset\: \text{and}\: S_C \neq \emptyset$}
    \State $M_F[k] = \text{mean}_L (S_F, S_{\Delta f})$, $M_C[k] = \text{mean}_L (S_{C}, S_{\Delta f})$; \label{algline:shade-update-history}
    \State $k = (k+1) \; \text{modulo} \; H$; \label{algline:shade-k-increment-modulo}
\EndIf
  }
\State $t = t+1$; \label{algline:shade-last}
\EndWhile
\end{algorithmic}
\end{algorithm}

The success history is stored in arrays
$M_F=(M_F[1],\cdots,M_F[H])$ and $M_C=(M_C[1],\cdots,M_C[H])$, both initialized at the beginning of the search (line~\ref{algline:shade-init-histories}) to 0.5,  midway in the typical range [0, 1] for $F$ and $C$.

During the search, for each individual $x^{i,t}$, an integer $r^i$ is selected uniformly at random from the interval [1,H]. The corresponding elements $M_F[r^i]$ and $M_C[r^i]$ are used to generate $F^{i,t}$ and $C^{i,t}$, respectively (line~\ref{algline:shade-choose-history-index}). 
$F^{i,t}$ is drawn from a Cauchy distribution centered at $M_F[r^i]$ with scale $\gamma=0.1$, and $C^{i,t}$ is drawn from a normal distribution centered at $M_C[r^i]$ with standard deviation $\sigma =0.1$ (lines~\ref{algline:shade-generate-f}--\ref{algline:shade-generate-c}).
The sampling is repeated until a valid value ($>0$) for $F^{i,t}$ is obtained. The use of a Cauchy distribution, which tends to generate larger values compared to the normal distribution, has been found effective in benchmark problems.

If the offspring $u^{i,t}$ has a fitness equal to or better than its parent, the values $F^{i,t}, C^{i,t}$, and the fitness improvement $f(x^{i,t}) - f(u^{i,t})$ are stored in sets $S_F, S_{C}, S_{\Delta f}$, respectively (line~\ref{algline:shade-add-to-sets}). These sets are referred to as the {\it successful parameters}.

After all offspring have been generated and evaluated, the $k$-th elements of $M_F$ and $M_C$ are updated using a weighted Lehmer mean, with $\Delta f$ as the weight (line~\ref{algline:shade-update-history}). The Lehmer mean of values $X_i (i=1,...,n)$ with weights $w_i$ is defined as:
$\text{mean}_L(X, w) = \sum{(w_i\cdot X_i^2)} /  \sum{(w_i\cdot X_i)}$
The index $k \in \{1,...,H\}$ is initialized to 1 (line~\ref{algline:shade-init-k}) and incremented cyclically (modulo $H$) with each success-history update (line~\ref{algline:shade-k-increment-modulo}). The Lehmer mean, which tends to produce larger averages than the arithmetic mean, has been shown to be more effective for this purpose.
In the reference implementation by the SHADE authors~\cite{shadecode}, the population size is set to$|P^t|=100$, the history length to $H=D$, and $x^{\text{pbest},t}$ is selected from the top 10\% of the population ranked by fitness.

\paragraph{Mutation with Archive} \label{sec:archive}

SHADE also incorporates an archive mechanism, first introduced in JADE~\cite{ZhangS09}. Empirical results show that JADE with an archive achieves superior fitness on many 100-dimensional benchmark problems compared to its archive-free counterpart~\cite{ZhangS09}. Archives tend to improve performance in high-dimensional ($\geq 50$) problems and have become standard in many DE variants since JADE, including SHADE~\cite{TanabeF13}, often in conjunction with the current-to-$pbest$ mutation strategy.

The use of the archive in SHADE is highlighted in red in Algorithm~\ref{alg:SHADE}. The archive $A^t$ is initially empty (line~\ref{algline:SHADE-archive}) and accepts parents $x^{i,t}$ replaced by offspring during selection until the archive reaches its maximum size $|A|_\text{max}$ (line~\ref{algline:SHADE-add1}). Once full, a random individual in $|A^t|$ is replaced by $x^{i,t}$ (lines~\ref{algline:SHADE-add2}--\ref{algline:SHADE-add3}).

During mutation, SHADE selects $x^{r2}$ uniformly from $P^t \cup A^t$ (line~\ref{algline:SHADE-mutation}).
If $x^{r2}$ is from $A^t$, the difference vector $F\cdot(x^{r1}-x^{r2})$ becomes a vector from a previously removed individual toward a current individual. As the population converges toward promising regions, this vector promotes exploration and helps maintain diversity. JADE uses an archive size of $|A|_\text{max}=|P^t|$, while the SHADE reference implementation~\cite{shadecode} uses $|A|_\text{max}=2\cdot|P^t|$. In general, archive sizes are set larger than the population size.

\subsubsection{Linear population size reduction (LSHADE)} \label{sec:LPSR}

LSHADE~\cite{TanabeF14CEC} is an improvement to SHADE which incorporates a mechanism for controlling the population size $|P^t|$.
LSHADE starts with a larger population and reduces it according to a linearly decreasing schedule, so that by the termination condition (either maximum evaluations budgets or time limit), the population size reaches its minimum ($|P|_\text{min}=4$). This allows the algorithm to concentrate its search resources on refining the best solutions near the end of the search. Given the initial population size $|P^{1}|$ and the maximum evaluation budgets $L^\text{evaluation}_\text{max}$, the next generation's population size $|P^{t+1}|$ is determined by:
\begin{eqnarray}
|P^{t+1}| = \text{round}( (|P^1| - 4) \cdot (1 - \sum^{t}_{j=1}{|P^j|} / L^\text{evaluation}_\text{max}) )+ 4 \label{eq:LPSR}
\end{eqnarray}
The worst-performing individuals are iteratively removed from the current population until the desired size $|P^{t+1}|$ is reached. This method is referred to as the {\it linear population size reduction strategy} (LPSR).
Note that in contrast to the success-history based adaptative control of $F$ and $C$ which is applied at a per-individual level, LPSR is a deterministic control of population size applied to the whole population.

LSHADE has been shown to outperform SHADE with a fixed population size~\cite{TanabeF14CEC}, and  LSHADE ranked first among 16 algorithms in the CEC2014 Real Parameter Single Objective Optimization Competition.
Many subsequent state-of-the-art DE variants, including most of the best-performing DE entries in the CEC Single Objective Optimization Competition series (2014-2022) have been based on LSHADE. Therefore, we adopt LSHADE as the baseline state-of-the-art DE for comparison in this paper.

%% file: recent_de.tex
\subsection{Related Work} 
\paragraph{Deterministic Control of Population Size}
Two broad approaches to controlling the population size in DE have been proposed. The first is population size reduction.
The prevalent approach is deterministic reduction, using fixed (problem-independent) schedule for population size reduction.
Consequently, once the population has been reduced, there is no mechanism to increase it again, making it difficult to escape stagnation caused by insufficient diversity.
Nevertheless, the Linear Population Size Reduction (LPSR) scheme described above in Section \ref{sec:LPSR} has been widely adopted due to its effectiveness and simplicity.

Another deterministic approach is in dynNP-DE algorithm~\cite{brest2008population}, which  halves the population size repeatedly over time. Unlike LSHADE, which removes the worst-performing individuals based on fitness rankings, dynNP-DE deletes the less fit of two randomly selected individuals, effectively implementing a single-round tournament selection. EsDEr-NR~\cite{awad2018ensemble} reduces individuals in the later stages of evolution by removing those located near low-performing niches—regions far from the current best solution. Similarly, Distributed DE with Explorative-Exploitative Population Families (DDE-EEPF)~\cite{weber2009distributed} reduces the size of one of its subpopulations.

The second approach to population size control incorporates mechanisms to {\it increase} the population size when stagnation is detected. Unlike reduction, increasing the population size is non-trivial, and various distinct strategies have been proposed. For example, DEVP~\cite{wang2011adaptive} adds new individuals by mutating the current best solution when the best-so-far fitness ceases to improve. DEWAPI~\cite{elsayed2013differential} includes an intermediate evolution phase before incorporating new individuals into the main population. APTS~\cite{ZhuTFZ13} perturbs better-fitness individuals to generate new candidates. Guo et al.~\cite{guo2014improving} propose reintroducing archived individuals into the main population to increase size and promote exploration.

To facilitate population size adaptation, various indicators of search stagnation have also been proposed. For instance, SapsDE~\cite{wang2013differential} first reduces the population upon detecting stagnation and increases it only if stagnation persists. Population-Entropy SHADE (PESHADE)~\cite{zhang2021enhancing} uses population density as a metric, reducing density in the early stages to promote exploration and increasing it later to intensify search around promising regions. Polakova et al.~\cite{polakova2019differential} employ a diversity indicator $V$, adjusting the population when $V$ deviates from a scheduled linear decay curve, bounded within 90\% to 110\% of the expected value.

While population reduction methods are relatively simple, there are two issues: First, deterministic, problem-independent schedules may not be aligned with the problem characteristics. If the reduction is too aggressive, the search may stagnate; if too slow, promising regions may not be explored thoroughly. Second, deterministic schedules assume that  evaluation budgets (e.g., time or number of evaluations) are determined a priori -- in many applications, more flexibility is required (e.g., we may want to terminate the search earlier or later than assumed by the standard schedule).

Mechanisms for increasing population size are even more difficult to implement successfully -- they require not only reduction strategies but also stagnation detection and population augmentation processes, resulting in more complex control logic. A notable attempt at adaptive control without explicit indicators is DESAP~\cite{Teo06}, which uses self-adaptation: each individual maintains a population size parameter, and the average of these values determines the size of the next generation.
This work introduces a novel third approach to population size control that differs from the existing two strategies.

\paragraph{The Archive as a Secondary Population}
Section~\ref{sec:archive} introduced the use of an archive in the current-to-$pbest$ mutation operator~\cite{ZhangS09}. Beyond this, several DE variants employ a auxiliary population, in addition to the main population $P^t$, from which parent individuals are selected. We use the term {\it external archive} to refer to all such auxiliary populations, including the original archive in ~\cite{ZhangS09}.

For example, DESSA~\cite{lu2014new} maintains an external archive to construct surrogate models approximating the objective function. Gonuguntla et al.~\cite{gonuguntla2015differential} employ a "huge set" $S^t$, from which a subset $A^t$ is drawn each generation for use in search. When the average fitness of$S^t$ exceeds its median, indicating concentration near a local optimum, $A^t$ is composed of randomly selected individuals from $S^t$ to promote exploration. Otherwise, tournament selection is applied to form $A^t$, focusing the search near fitter individuals.
Guo et al.\cite{guo2014improving} propose the Successful-Parent-Selecting method, which reinserts archived individuals into the population upon detection of stagnation. Epitropakis et al.\cite{epitropakis2013dynamic} introduce a niching DE inspired by Particle Swarm Optimization (PSO), incorporating a dynamic archive~\cite{zhai2011dynamic}. In this method, if an offspring lies close to an archived individual, it is regenerated to maintain diversity. Other approaches, such as $\epsilon$-MyDE~\cite{santana2005algorithm} and~\cite{chang1999pareto}, utilize Pareto-optimal sets to preserve better-fitness solutions.
These external archive mechanisms highlight the need to overcome the limitations of relying solely on a single, fixed-size population maintained through generational replacement.

\paragraph{Incorporating Selection Pressure into Mutation} \label{sec:pressure}

In conventional rand/1 and current-to-$pbest$ mutation strategies, individuals such as $x^{r1}, x^{r2}, x^{r3}$ are typically selected uniformly at random. Selection pressure refers to the bias introduced to favor the selection of better-fitness individuals. When applied to mutation, this pressure increases the likelihood that fitter individuals are selected as components of mutant vectors.

Several mechanisms exist for introducing selection pressure into mutation, with fitness-proportional selection, rank-based selection, and tournament selection being the most common.
Empirical studies such as~\cite{stanovov2019selective} have shown that applying selection pressure (e.g., via fitness-proportional, rank-based~\cite{whitley1989genitor}, or tournament selection~\cite{goldberg1991comparative, miller1995genetic}) improves performance relative to uniform random selection in several mutation strategies.

Sutton et al.~\cite{SuttonLW07} demonstrated that selecting the individuals $x^{r1,t}, x^{r2,t}, x^{r3,t}$ used in the rand/1 mutation strategy (Equation~\ref{eq:rand1}) based on fitness-ranked selection significantly improves the performance of DE, particularly when the population size is large. 
ISSDE~\cite{lou2012differential} adopts a strategy in which individuals are selected to satisfy $f(x^{r1,t}) \leq f(x^{r2,t}) \leq f(x^{r3,t})$. This approach is equivalent to employing tournament selection with a tournament size of $T=3$ to determine $x^{r1,t}$.
Gong et al.~\cite{gong2013differential} showed that applying rank-based selection to $x^{r1,t}, x^{r2,t}$ within both the rand/1 and current-to-pbest mutation strategies (Equation~\ref{eq:cur-to-pbest}) can improve the performance of DE variants such as JADE~\cite{ZhangS09}.
LSHADE-RSP~\cite{stanovov2018lshadersp} %
employs rank selection~\cite{jebari2013selection} within the current-to-pbest mutation strategy.

\paragraph{On the ubiquity of archives and population size reduction in state-of-the-art DE}

Table~\ref{tab:adaptive-DE} surveys the use of success-history adaptation, population size reduction strategies, and archive usage in adaptive DE algorithms developed since JADE, specifically, 
all DE variants that ranked highly in the CEC Real-Parameter Single-Objective Optimization Competitions from 2014 to 2022~\cite{cec2014, cec2015, cec2016, cec2017, cec2018, cec2019, cec2020, cec2021, cec2022}. %

Table~\ref{tab:adaptive-DE} shows the prevalence of 3 components in state-of-the-art DE algorithms since LSHADE: (1) success-history adaptation, (2) population size reduction, and (3) archives.
The consistent usage of these three comonents  in top-ranking entries in the CEC Real-Parameter Single-Objective Optimization Competitions suggests a widely shared recognition that they are critical for enhancing the search performance of DE algorithms.
\footnote{The jDE100~\cite{brest2019100} and its variants do not have an archive, but use two populations with different roles to conduct the search.}

\input appendix_adaptiveDE

\paragraph{Evolutionary Computation with Infinite Populations}
Previous work has used infinite populations to facilitate the theoretical analysis of evolutionary algorithms.
The analysis of schema in genetic algorithms assumes an infinite population \cite{Holland75}.
Qin et al. analyze genetic algorithms~\cite{qi1994theoreticalP1},\cite{qi1994theoreticalP2} by assuming an infinite number of populations,
By assuming an infinite population, the authors derive how the population's mean positions and correlations among its coordinates evolve by mutation and crossover.
A similar analysis was also performed by Whitley et al.~\cite{whitley1993executable} and tested on a GA with a population size of 625 in a 15-dimensional problem.

In these works, infinite populations are used in  order to simplify theoretical analysis. Applying these models to predict/understand the behavior of actual, finite populations is challenging, but they may provide insights into search behavior in the limit as population size increases.
In contrast, we consider the use of monotonically increasing, finite populations with unbounded size as a practical alternative to search algorithms with standard (fixed size with generational replacement) population structures.

%% file: appendix_adaptiveDE.tex
\footnotesize

\begin{table} %
\caption{\small
Success-history (SH), population size reduction (PSR) strategies and archive use in
the IEEE CEC Real-Parameter Single-Objective Optimization Competition series (RPSOOC 2014-2022)
~\cite{cec2014, cec2015, cec2016, cec2017, cec2018, cec2019, cec2020, cec2021, cec2022}.
}
\label{tab:adaptive-DE} 
\begin{tabular}[c]{lcccccc}
name & reference & year &  SH & archive &  PSR & remarks \\\hline\hline

JADE & \cite{ZhangS09} & 2009 &  & $\surd$ &  &Proposed widely used archive method \\
SHADE & \cite{TanabeF13} & 2013 & $\surd$ & $\surd$ &  & CEC2013 RPSOOC 4th \\ %
LSHADE & \cite{TanabeF14CEC} & 2014 & $\surd$ & $\surd$ & $\surd$ & CEC2014 RPSOOC 1st\\ %
SPS-LSHADE-EIG & \cite{guo2015self} & 2015 & $\surd$ & $\surd$ & $\surd$ & CEC2015 RPSOOC 1st\\
LSHADE-EpSin & \cite{awad2016ensemble} & 2016 & $\surd$ & $\surd$ & $\surd$  & CEC2016 RPSOOC 2nd\\ %
iLSHADE & \cite{brest2016shade} & 2016 & $\surd$ & $\surd$ & $\surd$ & CEC2016 RPSOOC 3rd\\% &
jSO & \cite{brest2017single} & 2017 & $\surd$ & $\surd$ & $\surd$ & CEC2017 RPSOOC 2nd\\ %
LSHADE-cnEpSin & \cite{awad2017ensemble} & 2017 & $\surd$ & $\surd$ & $\surd$ & CEC2017 RPSOOC 3rd\\ %
LSHADE-SPACMA & \cite{mohamed2017lshade} & 2017 & $\surd$ & $\surd$ & $\surd$ & CEC2017 RPSOOC 4th\\ %
LSHADE-RSP & \cite{stanovov2018lshadersp} & 2018 & $\surd$ & $\surd$ & $\surd$ & CEC2018 RPSOOC 2nd\\ %
ELSHADE-SPACMA & \cite{hadi2021single} & 2018 & $\surd$ & $\surd$ & $\surd$ & CEC2018 RPSOOC 3rd\\ %
jDE100 & \cite{brest2019100} & 2019 & & &  & CEC2019 RPSOOC 1st\\
DISHchain1e+12 & \cite{zamuda2019function} & 2019 & $\surd$ & $\surd$ & $\surd$ & CEC2019 RPSOOC 2nd\\ 
IMODE & \cite{sallam2020improved} & 2020 & $\surd$ & $\surd$ & $\surd$ & CEC2020 RPSOOC 1st\\ %
AGSK & \cite{mohamed2020evaluating} & 2020 & & & $\surd$  & CEC2020 RPSOOC 2nd\\
j2020 & \cite{brest2020differential} & 2020 & & & & CEC2020 RPSOOC 3rd\\
OLSHADE & \cite{biswas2020large} & 2020 & $\surd$ & $\surd$ & $\surd$  & CEC2020 RPSOOC 4th\\ %
jDE100e & \cite{bujok2020eigenvector} & 2020 & & & & CEC2020 RPSOOC 5th\\
NL-SHADE-RSP & \cite{stanovov2021nl} & 2021 & $\surd$ & $\surd$ & $\surd$  & \begin{tabular}{c}1st in two CEC2021 RPSOOC events\\ and 3rd in one RPSOC event
\end{tabular}\\ %
jDE21 & \cite{brest2021self} & 2021 & & & $\surd$ & \begin{tabular}{c}
1st in one CEC2021 RPSOOC event\\ and 2rd in two RPSOC events\end{tabular}\\
APGSK-IMODE & \cite{mohamed2021gaining} & 2021 & $\surd$ & $\surd$ & $\surd$  & \begin{tabular}{c}1st in one CEC2021 RPSOOC event\\ and 4th in three RPSOC events
\end{tabular}\\ %
DEDMNA & \cite{bujok2021differential} & 2021 & $\surd$ & $\surd$ & $\surd$ & \begin{tabular}{c}1st in one CEC2021 RPSOOC event\\ and 3rd in two RPSOC events\end{tabular}\\ %
EA4eig & \cite{bujok2022eigen} & 2022 & $\surd$ & $\surd$ & $\surd$  & CEC2022 RPSOOC 1st\\ %
NL-SHADE-LBC & \cite{stanovov2022nl} & 2022 & $\surd$ & $\surd$ & $\surd$  & CEC2022 RPSOOC 2nd\\ %
NL-SHADE-RSP-MID & \cite{biedrzycki2022version} & 2022 & $\surd$ & $\surd$ & $\surd$  & CEC2022 RPSOOC 3rd\\ %
S-LSHADE-DP & \cite{van2022dynamic} & 2022 & $\surd$ & $\surd$ & $\surd$  & CEC2022 RPSOOC 4th\\ %
jSObeE & \cite{kolenovsky2022adaptive} & 2022 & $\surd$ & $\surd$ & $\surd$  & CEC2022 RPSOOC 5th\\
\hline
\end{tabular}
\end{table}

\normalsize

%% file: ude-jp.tex
\subsection{Differential Evolution without Generational Replacement: Unbounded DE}
\label{sec:ude}
In Differential Evolution (DE), generational replacement is believed to play a crucial role in preserving better-fitness individuals and focusing the search on promising regions of the solution space. Nevertheless, prior studies have shown that DE variants which retain discarded individuals in an auxiliary archive population outperform traditional single-population DE methods~\cite{ZhangS09, gonuguntla2015differential, guo2014improving}.
This suggests that individuals dicarded by generational replacement may still possess valuable information for search.
However, the use of an archive requires careful design choices, including parameters such as retention duration, criteria for inclusion and deletion, and other operational rules.

 Motivated by these observation, we propose Unbounded Differential Evolution (UDE), a DE variant that operates on a single population and retains all individuals, including those superseded by offspring. In UDE, even when an offspring is successful, its parent is not discarded. Unlike theoretical models that consider infinite population sizes for analytical convenience~\cite{whitley1993executable, qi1994theoreticalP1, qi1994theoreticalP2}, UDE is a practical algorithm with an unbounded, but not infinite, population size.

\paragraph{Advantages of UDE}
Although generational replacement is traditionally regarded as a key component of DE, UDE avoids discarding potentially useful individuals.
As such, it provides a simple and unified framework which enables the simulation  of having a primary population and auxiliary population (archive)  within a single, unbounded population. Additionally, in the UDE framework,  {\it selective pressure}, rather than generational replacement, drives the selection process. While some recent DE variants have incorporated selective pressure in a limited fashion \cite{stanovov2018lshadersp,stanovov2021nl,biedrzycki2022version,stanovov2022nl}, UDE extensively applies selective pressure for all selection operations.  Thus, UDE is a conceptually simple, significant departure from standard DE.

Furthermore, by applying adaptive parameter control to the selection policy itself, UDE can implement search behavior similar to that of adaptive control of population size -- a feature traditionally associated with increased algorithmic complexity.
In UDE, the computational cost for individual selection increases as the population size grows. However, in most practical applications of evolutionary computation, the time required to generate individuals are negligible compared to the computational cost of evaluating objective functions, and thus this overhead is typically not a concern.

Unbounded Success-history based Adaptive DE (USHADE) is intended to exemplify DE with adaptive parameter control, and extends UDE by incorporating the widely used, success-history based adaptive parameter control mechanism from SHADE.

\subsubsection{UDE} \label{sub:ude} %
Algorithm~\ref{alg:uDE} presents the pseudocode for UDE. While the algorithm shown employs the current-to-$pbest$ mutation strategy—commonly used in state-of-the-art DE variants, UDE is a general framework, and other  strategies, such as rand/1, can be adopted by modifying lines~\ref{algline:ude-selection} and~\ref{algline:ude-mutation}.

\begin{algorithm}[htbp]
\caption{UDE (\textcolor{red}{red} is difference from baseline-DE).}
\label{alg:uDE}
\begin{algorithmic}[1]
\State $t = 1$;
\State Initialize population $P^t$; 
\For{ $i = 1$ to $|P^1|$}
    \State evaluate $f(x^{i,t})$;
\EndFor
\While{not termination condition}
\For{ $i = 1$ to $gensize$} \label{algline:ude-gen}
    \State select $x^{pbest,t}$
    \State \textcolor{red}{select $x^{p,t}, x^{r1,t}, x^{r2,t}$ from $P^t$ with selection policy $(p \neq r1, p \neq r2, r1\neq r2)$;} \label{algline:ude-selection}
    \State generate $v^{i,t}$ using current-to-$pbest$ mutation; \label{algline:ude-mutation}
    \State generate $u^{i,t}$ using binomial crossover between $v^{i,t}$ and $x^{p,t}$; \label{algline:ude-crossover}
\EndFor
\For{ $i = 1$ to $gensize$} \label{algline:ude-eval}
    \State evaluate $f(u^{i,t})$;
    \State \textcolor{red}{$P^{t+1} = P^t \cup u^{i,t}$;} \label{algline:ude-success}
\EndFor
\State $t = t+1$;
\EndWhile
\end{algorithmic}
\end{algorithm}

\begin{figure}[htbp]
\begin{center}
\includegraphics[width=0.9\textwidth]{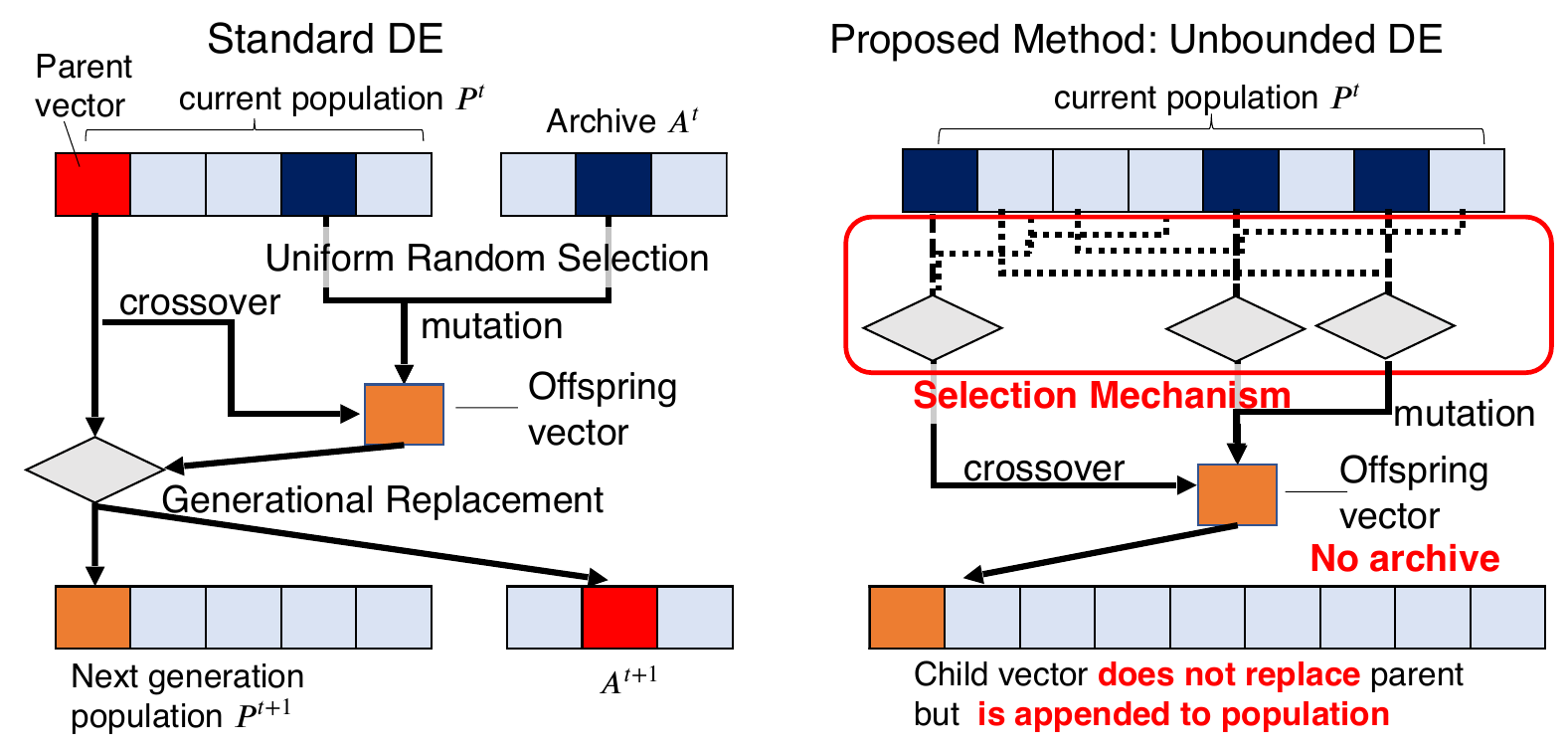}
\end{center}
\caption{\small
Comparison of conventional DE (left) vs. UDE (right). Boxes represent populations, and arrows indicate mutation, crossover, and population update. The diamond shape indicates selection by comparison of fitness.
}
\label{fig:abs}
\end{figure}

Figure~\ref{fig:abs} shows a conceptual diagram of conventional DE and UDE.
When compared to conventional DE employing the current-to-$pbest$ mutation strategy, we call this baseline-DE, UDE differs in three key aspects:
\begin{enumerate}
\item Population growth (line~\ref{algline:ude-success}): Offspring are added to the population without replacing their parents, resulting in a monotonically increasing population size.
\item Selection of parent $x^{p,t}$ (line~\ref{algline:ude-selection}): In standard DE,  each individual serves as a parent exactly once per generation %
  In contrast, UDE selects parent individuals dynamically.
\item Definition of ``generation'' (line~\ref{algline:ude-gen} and \ref{algline:ude-eval}): In standard DE, a generation is the creation/evaluation of $|P^t|$  new individuals, but in UDE, a ``generation'' is the creation/evaluation of $gensize$ individuals, where $gensize$ is constant throughout the search.
  
\end{enumerate}

\subsubsection{USHADE} \label{sub:uade} %
Unbounded Success-history based Adaptive DE (USHADE) is an extension of UDE that incorporates adaptive parameter control. As described in Section~\ref{sec:parameter}, state-of-the-art DE algorithms typically employ adaptive control for parameters such as the scaling factor $F$ and crossover rate $C$. USHADE applies the same success-history based adaptation scheme as SHADE.
Algorithm~\ref{alg:uADE} presents the pseudocode for USHADE.
In addition to the differences between UDE and DE which were already described above in Section \ref{sub:ude}, 
a major  difference between USHADE and SHADE (Algorithm~\ref{alg:SHADE}) is 
elimination of the external archive (highlighted in red in Algorithm~\ref{alg:SHADE}): There is no need for an explicit, auxiliary population because all individuals, including those which would traditionally be stored in an archive, are retained in $P^t$, and the decision of whether to use newer or older individuals is left up to the selection policy.

\subsection{Selection Policy} \label{sub:selection-policy} %
The selection policy governs the choice of all individuals required for variation, including the parent $x^{p,t}$, as well as auxiliary individuals such as $x^{r1,t}$ and $x^{r2,t}$ used in the current-to-$pbest$ mutation strategy (see line~\ref{algline:ude-selection} in Algorithm~\ref{alg:uDE} and line~\ref{algline:uADE-selection} in Algorithm~\ref{alg:uADE}). While UDE is compatible with various mutation strategies, this paper adopts the widely used current-to-$pbest$ mutation, in line with state-of-the-art DE practices. For other mutation strategies, the selection policy should be appropriately adapted to select the required individuals.
For example, in the case of rand/1 mutation (Equation~\ref{eq:rand1}), the selection policy would need to choose four individuals: 
$x^{p,t}, x^{r1,t}, x^{r2,t}$ and $x^{r3,t}$. %

\begin{algorithm}[htbp]
\caption{USHADE (\textcolor{red}{red} is difference from Alg.~\ref{alg:SHADE} SHADE).}
\label{alg:uADE}
\begin{algorithmic}[1]\small
\State $t = 1$;
\State Initialize population $P^t$;
\State Initialize contents of success histories $M_F$ and $M_C$ to 0.5; 
\State $k = 1$;
\For{ $i = 1$ to $|P^t|$}
    \State evaluate $f(x^{i,t})$;
\EndFor
\While{not termination condition}
\State $S_F = \emptyset, S_{C} = \emptyset$;
\For{ $i = 1$ to $gensize$}
    \State select $r^i$ randomly from $\{1,\cdots,H\}$;
    \While{$F^{i,t}\leq0$}
        \State $F^{i,t}=\text{min}(\text{rand}_\text{cauchy}((M_F[r^i], 0.1),1)$; \label{algline:uADE-generate-f}
    \EndWhile
    \State $C^{i,t}=\text{max}(0,\text{min}(\text{rand}_\text{normal}(M_C[r^i], 0.1),1))$; \label{algline:uADE-generate}
    \State select $x^{pbest,t}$
    \State \textcolor{red}{select $x^{p,t}, x^{r1,t}, x^{r2,t}$ from $P^t$ with selection policy $(p \neq r1, p \neq r2, r1\neq r2)$;} \label{algline:uADE-selection}
    \State generate $v^{i,t}$ using current-to-{\it pbest} mutation; 
    \State generate $u^{i,t}$ using binomial crossover between $v^{i,t}$ and $x^{p,t}$;
\EndFor
\For{ $i = 1$ to $gensize$}
    \State evaluate $f(u^{i,t})$;
    \State \textcolor{red}{$P^{t+1} = P^t \cup u^{i,t}$;} \label{algline:uADE-addition}
    \If{$f(u^{i,t}) \leq f(x^{p,t})$} \label{algline:uADE-replacement}
        \State $S_F = S_F \cup F^{i,t}$, $S_{C} = S_{C} \cup C^{i,t}$, $S_{\Delta f} = S_{\Delta f} \cup (f(x^{p,t}) - f(u^{i,t}))$; \label{algline:uADE-add-to-sets}
    \EndIf
\EndFor
\If{$S_F \neq \emptyset\: \text{and}\: S_C \neq \emptyset$}
    \State $M_F[k] = \text{mean}_L (S_F, S_{\Delta f})$, $M_C[k] = \text{mean}_L (S_{C}, S_{\Delta f})$;\label{algline:uADE-update-history}
    \State $k = (k+1) \; \text{modulo} \; H$;
\EndIf
\State $t = t+1$;
\EndWhile
\end{algorithmic}
\end{algorithm}

The simplest selection policy involves selecting individuals uniformly from the population $P^t$. However, selecting all individuals (old and new) hampers the algorithm's ability to focus exploration on promising regions of the search space.

One effective approach to bias offspring generation toward promising regions is to employ tournament selection \cite{goldberg1991comparative, miller1995genetic}.
Tournament selection biases selection so that better individuals are more likely to serve as parents in offspring generation.
We adopt a tournament selection scheme which samples a subset of $n \geq 1$ individuals uniformly at random from the population and selects the individual with the best fitness from this subset. The probability that an individual of fitness rank $i$-th is selected (assuming the population is sorted by fitness) can be expressed using the combination formula ${}_n\mathrm{C}_r$. Increasing $n$ intensifies the selection pressure, favoring better-fitness individuals, while decreasing $n$ allows for more diverse, potentially lower-fitness individuals to be selected.
\[
p(i) = 
\begin{cases}
{
\frac{{}_{|P^t|-i}\mathrm{C}_{n-1}} {{}_{|P^t|}\mathrm{C}_n}\; \text{if} \: i \in [1,|P^t|-n-1]
} \\
{
0\; \text{if} \: i \in [|P^t|-n, |P^t|]
}
\end{cases}
\]

As a baseline tournament policy, we define the T policy,
which select $n$ ($=|P^t| / T$) individuals uniformly at random from the population $P^t$ to obtain a candidate pool, and chooses the individual with the best fitness is chosen.

The value of $T$ is fixed to $|P^1|$ in UDE, and in USHADE, $T^{i,t}$ is controlled adaptively as described below in Section \ref{sec:adaptive-selection-policy-parameter}.  %

\subsubsection{Diversity-Preserving Tournament Policy} \label{sec:ex1}

We now introducse a tournament policy designed to maintain greater diversity than the baseline T policy.

The {\it Diversity-Preserving Tournament} (DPT) policy is as follows: 
For each offspring to be generated (totaling $gensize$), select an integer $j$ between 1 and $gensize$. Construct a subset $S \in P$ consisting of individuals whose insertion index modulo $gensize$ equals $j$. From $S$, conduct uniform sampling to obtain $n$ ($=|P^t| / T$) candidates, and select the best among them based on fitness.
Given that $gensize$ is fixed, the individuals in $S$ are expected to be related through parent-offspring lineage. When generating offspring $x^{i,t}$, the respective index values $j^p, j^{r1}, j^{r2}$ for selecting the parent $x^p$ and two additional $x^{r1}, x^{r2}$ individuals are determined as follows:
\begin{align}
j^p &= i \\[-3pt]
j^{r1} &= \text{rand}_\text{uniform} (1,\cdots,gensize), j^{r1} \neq j^{p} \\[-3pt]
j^{r2} &= \text{rand}_\text{uniform} (1,\cdots,gensize), j^{r2} \neq j^{p}, j^{r2} \neq j^{r1}
\end{align}

The purpose of tournament selection is to preferentially choose better-fitness individuals while preserving a low probability of selecting lower-quality ones. Compared to T, the DPT policy is expected to maintain greater diversity, as it reduces the likelihood of repeatedly selecting elite individuals from the same subset. Moreover, when the crossover rate $C$ is small, parent and offspring relationships tend to share many variable values, increasing intra-subset similarity. By localizing tournaments to different subsets of the population for each offspring, DPT helps mitigate diversity loss.

\subsubsection{Adaptive Control of Tournament Selection Policy Parameter} %
\label{sec:adaptive-selection-policy-parameter}
We now describe a method for adaptively controlling the parameter $T^{i,t}$ used in the selection policies. The red-highlighted sections in Algorithm~\ref{alg:uPADE} show how $T^{i,t}$ is managed.
\input alg-uPADE.tex

The parameter $T^{i,t}$ is handled analogously to the scale factor $F$ and $C$. Specifically, a success-history $M_T$ is initialized with the value $|P^1|$ (line~\ref{algline:uPADE-initialize}). In each generation, $T^{i,t}$ is drawn from a normal distribution centered around a value from the success-history (line~\ref{algline:uPADE-generate-t}), similar to the approach for $C^{i,t}$. To account for the larger numerical scale of $T^{i,t}$, the standard deviation is set to $\sigma_T=10$, which is larger than $\sigma_{C}=0.1$ used for $C^{i,t}$. The lower bound of $T^{i,t}$ is fixed at 100, with no upper bound.
When an offspring is successful, $T^{i,t}$ is added to the set $S_T$ (line~\ref{algline:uPADE-add-to-sets}), and the relevant success-history element $M_T[k]$ is set to the Lehmer mean, weighted by the fitness improvement $S_{\Delta f}$ (line~\ref{algline:uPADE-update-history-t}).

The performance of this algorithm is evaluated in Section~\ref{sec4}, with a detailed analysis provided in Section~\ref{sec:ex2}.

\subsection{UDE/USHADE without Failed Individuals}
\label{sec:ude-without-failed-individuals}
A defining feature of UDE is that all individuals are kept in the population, and selection is responsible which individuals to user or ignore as search progresses.
In contrast, discarding failed individuals (which do not have better fitness values than their parent) is a a standard feature of DE with generational replacement.

By moving line 22 of Algorithm~\ref{alg:uPADE} inside the conditional block at line 23, failed offspring will be excluded from the population update, thereby ensuring that only successful individuals are retained in the population. %

We refer to this variant of USHADE which discards failed individuals as {\it USHADE/DF}, and the non-adaptive variant {\it USHADE/DF}.
The preliminary conference version of this study  \cite{kitamura2022differential} presented extensive experimental evaluations of  UDE/DF and USHADE/DF (``UDE'' and ``USHADE'' in the previous paper refers to UDE/DF and USHADE/DF, respectively).
We experimentally compare USHADE/DF to USHADE
 in Section~\ref{sec:cec22}.

%% file: alg-uPADE.tex
\begin{algorithm}[htbp]
\caption{USHADE(T) (\textcolor{red}{red} is related to adaptive control of prameter $T$).}
\label{alg:uPADE}
\begin{algorithmic}[1]\small
\State $t = 1$;
\State Initialize population $P^t$;
\State Initialize contents of success histories $M_F$ and $M_C$ to 0.5, \textcolor{red}{$M_T$ to $|P^1|$}; \label{algline:uPADE-initialize}
\State $k = 1$;
\For{ $i = 1$ to $|P^t|$}
    \State evaluate $f(x^{i,t})$;
\EndFor
\While{not termination condition}
\State $S_F = \emptyset, S_{C} = \emptyset$;
\State \textcolor{red}{$S_T = \emptyset$;}
\For{ $i = 1$ to $gensize$}
    \State select $r^i$ randomly from $\{1,\cdots,H\}$;
    \While{$F^{i,t}\leq0$}
        \State $F^{i,t}=\text{min}(\text{rand}_\text{cauchy}((M_F[r^i], 0.1),1)$; \label{algline:uPADE-generate-f}
    \EndWhile
    \State $C^{i,t}=\text{max}(0,\text{min}(\text{rand}_\text{normal}(M_C[r^i], 0.1),1))$; \label{algline:uPADE-generate}
    \State \textcolor{red}{$T^{i,t}=\text{max}(100,\text{rand}_\text{normal}(M_T[r^i] ,10))$}; \label{algline:uPADE-generate-t}
    \State select $x^{pbest,t}$
    \State select $x^{p,t}, x^{r1,t}, x^{r2,t}$ from $P^t$ with \textcolor{red}{tournament policy} $(p \neq r1, p \neq r2, r1\neq r2)$; \label{algline:uPADE-mutation}
    \State generate $v^{i,t}$ using current-to-{\it pbest} mutation; 
    \State generate $u^{i,t}$ using binomial crossover between $v^{i,t}$ and $x^{p,t}$;
\EndFor
\For{ $i = 1$ to $gensize$}
    \State evaluate $f(u^{i,t})$;
    \State $P^{t+1} = P^t \cup u^{i,t}$; \label{algline:uPADE-addition}
    \If{$f(u^{i,t}) \leq f(x^{p,t})$} \label{algline:uPADE-replacement}
        \State $S_F = S_F \cup F^{i,t}$, $S_{C} = S_{C} \cup C^{i,t}$, $S_{\Delta f} = S_{\Delta f} \cup (f(x^{p,t}) - f(u^{i,t}))$, \textcolor{red}{$S_T = S_T \cup T^{i,t}$}; \label{algline:uPADE-add-to-sets}
    \EndIf
\EndFor
\If{$S_F \neq \emptyset\: \text{and}\: S_C \neq \emptyset$}
    \State $M_F[k] = \text{mean}_L (S_F, S_{\Delta f})$, $M_C[k] = \text{mean}_L (S_{C}, S_{\Delta f})$, \textcolor{red}{$M_T[k] = \text{mean}_L (S_T, S_{\Delta f})$;} \label{algline:uPADE-update-history-t}
    \State $k = (k+1) \; \text{modulo} \; H$;
\EndIf
\State $t = t+1$;
\EndWhile
\end{algorithmic}
\end{algorithm}

%% file: appendix-1.tex
\newcommand{\FinalBestSoFar}{\text{fitness}_{\text{best-so-far},\text{final}}}

We experimentally evaluate UDE and its variants using standard benchmark problem sets used to evaluate DE.
In this chapter, we first introduce a method for comparing evolutionary computation algorithms (Sec.~\ref{sec:postprocess}) and then present the results of a performance comparison (Sec.~\ref{sec:cec22}).
Sec.~\ref{simulating-popsize} shows that UDE mimics population size control through the behavior of the parameter.

All experiments were run on a Apple M3 Pro CPU with 16GB RAM running macOS Sequoia. All algorithms were implemented in C++ (g++ with -O3 and -std=c++11 compilation flags).

\subsection{Evaluation Methodology} \label{sec:postprocess} %

We use 2 sets of benchmarks from the IEEE Congress on Evolutionary Computation (CEC) Single Objective Optimization Competition series, specifically:
\begin{itemize}
  \item
The CEC2014 benchmark suite \cite{cec2014}, consisting of 30 problems
($F1$ Rotated High Conditioned Elliptic Function,
$F2$ Rotated Bent Cigar Function,
$F3$ Rotated Discus Function,
$F4$ Shifted and Rotated (S\&R) Rosenbrock's Function,
$F5$ S\&R Ackley's Function,
$F6$ S\&R Weierstrass Function,
$F7$ S\&R Griewank's Function,
$F8$ Shifted Rastrigin's Function,
$F9$ S\&R Rastrigin's Function,
$F10$ Shifted Schwefel's Function,
$F11$ S\&R Schwefel's Function,
$F12$ S\&R Katsuura Function,
$F13$ S\&R HappyCat Function,
$F14$ S\&R HGBat Function,
$F15$ S\&R Expanded Griewank's plus Rosenbrock's Function,
$F16$ S\&R Expanded Scaffer's f6 Function,
$F17-F22$ Hybrid Functions,
$F23-F30$ Composition Fuctions
)
\item
The CEC2022 benchmark suite \cite{cec2022}, consisting of 12 problems
($F1$ S\&R Zakharov Function,
$F2$ S\&R Rosenbrock's Function,
$F3$ S\&R Expanded Schaffer's f6 Function,
$F4$ S\&R Non-Continuous Rastrigin Fuction,
$F5$ S\&R Levy Function,
$F6-F8$ Hybrid Functions,
$F9-F12$ Composition Functions).
\end{itemize}
In hybrid functions, the $D$ dimensions are  divided into 3-5 groups, and a different function from above is assigned to each group. %
A composite function is a problem in which the final output is a weighted sum of the outputs of the above functions without dividing the dimensionality of the problem. %

The main objective of this paper is to explore the viability of the UDE framework (whose key features are unbounded population and  no generational replacment) as an alternative to standard, generational-replacement based DE.
However, in order to assess the UDE as a practical framework in the context of state-of-the-art DE, we focus our evaluation on USHADE, which incorporates success-history based adaptive control of $F$, $C$, and the (new in UDE) $T$ parameters.
As USHADE is UDE with the control parameter adaptation of SHADE, our experiments primarily use the CEC2014 benchmark suite,  because it was the benchmark set on which LSHADE was tuned \cite{TanabeF14CEC}, and is also very similar to the CEC2013 benchmark set on which SHADE was tuned \cite{TanabeF13}.

\subsubsection{Empirical Cumluative Distribution Functions (ECDF)}
\label{sec:ecdf}

To concisely illustrate and compare the performance of multiple algorithms, we use Empirical Cumulative (Run-Time) Distribution Functions (ECDFs). ECDFs are widely adopted in evolutionary computation research as an informative performance indicator~\cite{hansen2006compilation, qin2008differential}. The ECDF is formally defined as follows: given a performance threshold $z$ and trial outcomes $z_i (1 \le i \le N)$, where $N$ denotes the number of independent runs (trials), we define:
\[
\text{ECDF}(z) = \frac{1}{N}\text{count}(z_i < z) \quad i = 1,\cdots,N
\]
\noindent where 
$\text{count}(z_i < z)$ denotes the number of runs for which $z_i$ is less than the threshold $z$. The ECDF value lies in the range [0,1], where 1 is the best (all runs achieve the performance threshold).

In this paper, the ECDF plots (e.g., Figure~\ref{fig:cec14})  use evaluation count as the horizontal axis (up to the maximum evaluation budget).
The vertical axis represents average attainment rate (fraction of runs reaching  given targets), based on whether the best-so-far fitness $\text{fitness}_\text{best-so-far}$ (best fitness value observed among all invidiausl in the search up to that point in the search) achieves specified thresholds.

The target threshold values $z$ depend on the figure, and also depend on the data.
For most of the figures, $z$ is set as follows:
Let $\FinalBestSoFar$ be the
$\text{fitness}_\text{best-so-far}$ score at the end of a trial.
Then $z$ is the median value of $\FinalBestSoFar$ among all of the $NumTrials \times k$ trials, where $k$ is the number of algorithms being evaluated in the figure.
For example, suppose we are evaluating algorithms $A1$ and $A2$, and $NumTrials=2$.
If $\FinalBestSoFar(A1, trial1) = 9$, 
$\FinalBestSoFar(A1, trial2) = 2$, 
$\FinalBestSoFar(A2, trial1) = 5$, 
$\FinalBestSoFar(A2, trial2) = 6$, then $z$ is set to 5.5 (the median of 9, 2, 5, 6).

A key advantage of this approach to setting the $z$ threshold values is that $z$ is automatically determined from the performance data ($z$ is not decided arbitrarily)
A curve that lies higher on the plot at a given evaluation count indicates that a larger proportion of trials have reached the target by that number of evaluations, indicating better search performance.
However, it is important to note: (1) the appearance of the figure depends on which algorithms are included in the comparison (as the $z$ values depend on the data), and (2) since the vertical axis reflects success rates, improvements in $\text{fitness}_\text{best-so-far}$ beyond meeting the $z$ threshold are not reflected in the graph.

\subsubsection{Statistical Tests}
To complement the ECDF-based analysis, we employ the Wilcoxon rank-sum test\cite{wilcoxon1992individual} for statistical comparison of the $\FinalBestSoFar$ across algorithms. The Wilcoxon test is a non-parametric method that tests the null hypothesis that two groups come from the same distribution.

The Wilcoxon test only considers the $\text{fitness}_\text{best-so-far}$ at a specific point (usually the end of the search), and does not consider how fast the search reached that fitness value. For example, if algorithm A reaches the optimum in 1,000 evaluations and algorithm B reaches it in 10,000 evaluations, the two algorithms are statistically indistinguishable by a Wilcoxon test applied at 10,000 evaluations.
Therefore, combining ECDF plots (which capture search behavior over time) with the Wilcoxon rank-sum test (which assesses statistical significance at termination) provides a more comprehensive performance analysis.

%% file: tab-DE.tex
\begin{table}[htbp]
  \caption{\small
Comparison of baseline-DE vs. UDE(DPT) 
(CEC 2014 benchmarks, 10,30,50 dimension, maximum evaluation budgets $=20,000 \times D$). Wilcoxon ranked
sum test(p = 0.05) results on $F1$ to $F30$ are shown.
(\textcolor{red}{+}: better than UDE(DPT), \textcolor{blue}{-}: worse than UDE(DPT), $\approx$: no significant difference)}
\label{tab:DE}
\begin{tabular}{c|ccc|ccc|ccc}
   & \multicolumn{3}{l}{D=10}                   & \multicolumn{3}{l}{D=30}                   & \multicolumn{3}{l}{D=50}                   \\\hline
   & \multicolumn{2}{l}{baseline-DE} & UDE(DPT)  & \multicolumn{2}{l}{baseline-DE} & UDE(DPT)  & \multicolumn{2}{l}{baseline-DE} & UDE(DPT)  \\\hline\hline
1  & 36.18613        &\textcolor{blue}{-}             & 20.54281 & 7283335         &\textcolor{blue}{-}             & 723106   & 29431309        &\textcolor{blue}{-}             & 3574833  \\\hline
2  & 0               &\textcolor{red}{+}             & 36.73471 & 0               & $\approx$        & 0        & 3425.534        &\textcolor{blue}{-}             & 2695.386 \\\hline
3  & 0               & $\approx$        & 0        & 0               & $\approx$        & 0        & 20973.1         &\textcolor{blue}{-}             & 11339.33 \\\hline
4  & 27.40885        &\textcolor{red}{+}             & 32.2551  & 88.09352        &\textcolor{red}{+}             & 100.4618 & 97.06719        & $\approx$        & 100.9507 \\\hline
5  & 19.0255         &\textcolor{blue}{-}             & 10.25118 & 20.87361        & $\approx$        & 20.88263 & 21.09946        & $\approx$        & 21.09487 \\\hline
6  & 0.00E+00        & $\approx$        & 0        & 0.123975        &\textcolor{blue}{-}             & 0        & 0.010175        & $\approx$        & 0        \\\hline
7  & 0.002831        &\textcolor{blue}{-}             & 0.000145 & 0               & $\approx$        & 0        & 0               &\textcolor{red}{+}             & 2.29E-08 \\\hline
8  & 0.00E+00        & $\approx$        & 0        & 84.77389        &\textcolor{blue}{-}             & 75.60346 & 224.1105        &\textcolor{blue}{-}             & 209.6879 \\\hline
9  & 9.421369        &\textcolor{blue}{-}             & 1.127176 & 142.7767        &\textcolor{blue}{-}             & 124.2151 & 307.7097        &\textcolor{blue}{-}             & 280.8689 \\\hline
10 & 62.9752         &\textcolor{blue}{-}             & 2.223157 & 3707.907        &\textcolor{blue}{-}             & 2595.386 & 9183.029        &\textcolor{blue}{-}             & 8422.8   \\\hline
11 & 600.0482        &\textcolor{blue}{-}             & 153.3611 & 6168.291        &\textcolor{blue}{-}             & 5632.668 & 12510.2         &\textcolor{blue}{-}             & 11910.79 \\\hline
12 & 0.582939        &\textcolor{blue}{-}             & 0.414775 & 1.958013        &\textcolor{blue}{-}             & 1.735049 & 3.004111        & $\approx$        & 2.982396 \\\hline
13 & 0.099508        &\textcolor{blue}{-}             & 0.071085 & 0.218465        &\textcolor{blue}{-}             & 0.152863 & 0.311309        &\textcolor{blue}{-}             & 0.251433 \\\hline
14 & 0.142746        &\textcolor{red}{+}             & 0.265666 & 0.250093        &\textcolor{red}{+}             & 0.279787 & 0.311161        &\textcolor{red}{+}             & 0.326982 \\\hline
15 & 1.328008        &\textcolor{blue}{-}             & 1.185064 & 13.15558        &\textcolor{blue}{-}             & 11.99339 & 28.01447        &\textcolor{blue}{-}             & 26.35871 \\\hline
16 & 1.903234        &\textcolor{blue}{-}             & 0.852413 & 12.03599        &\textcolor{blue}{-}             & 11.07068 & 21.78963        &\textcolor{blue}{-}             & 21.27776 \\\hline
17 & 8.545029        &\textcolor{blue}{-}             & 9.927231 & 225.6945        &\textcolor{blue}{-}             & 102.5081 & 2247.481        &\textcolor{blue}{-}             & 1870.149 \\\hline
18 & 0.656424        &\textcolor{red}{+}             & 1.051892 & 2682.016        &\textcolor{blue}{-}             & 2412.903 & 5828.276        &\textcolor{blue}{-}             & 2727.039 \\\hline
19 & 1.215526        & $\approx$        & 1.683887 & 110.2838        &\textcolor{red}{+}             & 2587.551 & 10425.02        &\textcolor{blue}{-}             & 490.33   \\\hline
20 & 1.463286        &\textcolor{blue}{-}             & 1.33863  & 1195.434        &\textcolor{blue}{-}             & 706.1719 & 9931.824        &\textcolor{blue}{-}             & 8437.368 \\\hline
21 & 0.423931        &\textcolor{blue}{-}             & 0.001892 & 33.69248        &\textcolor{blue}{-}             & 21.14349 & 77.11374        &\textcolor{blue}{-}             & 64.60026 \\\hline
22 & 0.417995        &\textcolor{blue}{-}             & 0.303739 & 25.31531        &\textcolor{blue}{-}             & 20.74168 & 30.16714        &\textcolor{blue}{-}             & 28.32219 \\\hline
23 & 329.4575        & $\approx$        & 329.4575 & 315.2441        &\textcolor{blue}{-}             & 315.2441 & 344.0045        &\textcolor{red}{+}             & 344.0045 \\\hline
24 & 112.8872        &\textcolor{blue}{-}             & 101.5051 & 223.5123        & $\approx$        & 223.7081 & 268.0168        &\textcolor{red}{+}             & 268.9755 \\\hline
25 & 135.2575        & $\approx$        & 145.0995 & 205.3525        &\textcolor{blue}{-}             & 203.632  & 212.432         &\textcolor{blue}{-}             & 209.6394 \\\hline
26 & 100.1046        &\textcolor{blue}{-}             & 100.0743 & 100.2174        &\textcolor{blue}{-}             & 100.1485 & 100.2963        &\textcolor{blue}{-}             & 114.0382 \\\hline
27 & 42.474          & $\approx$        & 95.59849 & 302.1944        &\textcolor{blue}{-}             & 300      & 307.2405        & $\approx$        & 314.6883 \\\hline
28 & 382.5869        &\textcolor{blue}{-}             & 379.9183 & 806.0123        &\textcolor{red}{+}             & 818.2773 & 1084.342        &\textcolor{red}{+}             & 1142.797 \\\hline
29 & 1603.526        & $\approx$        & 1676.034 & 4305368         &\textcolor{red}{+}             & 12842590 & 261486.2        &\textcolor{blue}{-}             & 231395.3 \\\hline
30 & 161.041         &\textcolor{blue}{-}             & 159.2856 & 222.3386        & $\approx$        & 220.8157 & 244.603         &\textcolor{blue}{-}             & 236.5378 \\\hline
\end{tabular}
\end{table}

%% file: tab-alladd-D10.tex
\begin{table}[htbp]
  \caption{\small
Comparison of LSHADE, SHADE, USHADE(T) vs. USHADE(DPT)
(CEC 2014 benchmarks, 10 dimension, maximum evaluation budgets $=200,000$). Wilcoxon ranked
sum test(p = 0.05) results on $F1$ to $F30$ are shown.
(\textcolor{red}{+}: better than USHADE(DPT), \textcolor{blue}{-}: worse than USHADE(DPT), $\approx$: no significant difference)
}
\label{tab:uadeT-D10}
\centering
\begin{tabular}{c|cc|cc|cc|c}
   & \multicolumn{2}{c|}{LSHADE} & \multicolumn{2}{c|}{SHADE} & \multicolumn{2}{c|}{USHADE(T)} & USHADE(DPT) \\\hline\hline
$F1 $ & 0.000        & $\approx$      & 0.000        & $\approx$     & 0.000         & $\approx$       & 0.000    \\\hline
$F2 $ & 0.000        & $\approx$      & 0.000        & $\approx$     & 0.000         & $\approx$       & 0.000    \\\hline
$F3 $ & 0.000        & $\approx$      & 0.000        & $\approx$     & 0.000         & $\approx$       & 0.000    \\\hline
$F4 $ & 28.471       & $\approx$      & 26.852       & \textcolor{red}{+}          & 30.688        & $\approx$       & 31.370   \\\hline
$F5 $ & 11.318       & \textcolor{red}{+}           & 16.094       & $\approx$     & 18.701        & \textcolor{blue}{-}            & 16.024   \\\hline
$F6 $ & 0.000        & $\approx$      & 0.018        & $\approx$     & 0.000         & $\approx$       & 0.000    \\\hline
$F7 $ & 0.000        & \textcolor{red}{+}           & 0.001        & \textcolor{blue}{-}          & 0.002         & $\approx$       & 0.001    \\\hline
$F8 $ & 0.000        & $\approx$      & 0.000        & $\approx$     & 0.000         & $\approx$       & 0.000    \\\hline
$F9 $ & 2.283        & $\approx$      & 2.434        & \textcolor{blue}{-}          & 4.448         & \textcolor{blue}{-}            & 2.283    \\\hline
$F10$ & 0.002        & \textcolor{red}{+}           & 0.004        & \textcolor{red}{+}          & 0.660         & \textcolor{blue}{-}            & 0.080    \\\hline
$F11$ & 23.970       & $\approx$      & 39.146       & \textcolor{blue}{-}          & 134.187       & \textcolor{blue}{-}            & 26.324   \\\hline
$F12$ & 0.043        & \textcolor{red}{+}           & 0.072        & \textcolor{blue}{-}          & 0.095         & \textcolor{blue}{-}            & 0.056    \\\hline
$F13$ & 0.038        & \textcolor{blue}{-}           & 0.060        & \textcolor{blue}{-}          & 0.016         & \textcolor{red}{+}            & 0.031    \\\hline
$F14$ & 0.068        & \textcolor{red}{+}           & 0.098        & \textcolor{red}{+}          & 0.165         & \textcolor{blue}{-}            & 0.136    \\\hline
$F15$ & 0.341        & \textcolor{red}{+}           & 0.390        & \textcolor{red}{+}          & 0.710         & \textcolor{blue}{-}            & 0.452    \\\hline
$F16$ & 1.112        & \textcolor{blue}{-}           & 1.356        & \textcolor{blue}{-}          & 0.883         & \textcolor{blue}{-}            & 0.668    \\\hline
$F17$ & 0.030        & \textcolor{blue}{-}           & 0.036        & \textcolor{blue}{-}          & 0.008         & \textcolor{red}{+}            & 0.016    \\\hline
$F18$ & 0.116        & \textcolor{blue}{-}           & 0.130        & \textcolor{blue}{-}          & 0.027         & \textcolor{blue}{-}            & 0.021    \\\hline
$F19$ & 0.823        & \textcolor{blue}{-}           & 1.661        & \textcolor{blue}{-}          & 0.068         & \textcolor{blue}{-}            & 0.033    \\\hline
$F20$ & 0.009        & $\approx$      & 0.010        & $\approx$     & 0.026         & \textcolor{blue}{-}            & 0.015    \\\hline
$F21$ & 0.272        & \textcolor{blue}{-}           & 0.565        & \textcolor{blue}{-}          & 1.203         & \textcolor{blue}{-}            & 0.009    \\\hline
$F22$ & 0.051        & $\approx$      & 0.086        & \textcolor{blue}{-}          & 0.270         & \textcolor{blue}{-}            & 0.058    \\\hline
$F23$ & 322.998      & $\approx$      & 329.457      & $\approx$     & 329.457       & $\approx$       & 329.457  \\\hline
$F24$ & 107.093      & $\approx$      & 108.156      & \textcolor{blue}{-}          & 111.263       & \textcolor{blue}{-}            & 107.201  \\\hline
$F25$ & 140.928      & $\approx$      & 138.292      & $\approx$     & 157.679       & $\approx$       & 150.421  \\\hline
$F26$ & 100.045      & \textcolor{blue}{-}           & 100.063      & \textcolor{blue}{-}          & 100.018       & \textcolor{red}{+}            & 100.032  \\\hline
$F27$ & 30.646       & \textcolor{blue}{-}           & 112.925      & \textcolor{blue}{-}          & 96.818        & \textcolor{blue}{-}            & 77.145   \\\hline
$F28$ & 389.255      & $\approx$      & 399.779      & $\approx$     & 388.108       & $\approx$       & 382.836  \\\hline
$F29$ & 547.682      & \textcolor{blue}{-}           & 599.904      & \textcolor{blue}{-}          & 392.547       & \textcolor{blue}{-}            & 375.292  \\\hline
$F30$ & 158.880      & \textcolor{blue}{-}           & 158.902      & \textcolor{blue}{-}          & 158.721       & $\approx$       & 158.701  \\\hline
\end{tabular}
\end{table}

%% file: tab-alladd-D30.tex
\begin{table}[htbp]
  \caption{\small
Comparison of LSHADE, SHADE, USHADE(T) vs. USHADE(DPT)
(CEC 2014 benchmarks, 30 dimension, maximum evaluation budgets $=600,000$). Wilcoxon ranked
sum test(p = 0.05) results on $F1$ to $F30$ are shown.
(\textcolor{red}{+}: better than USHADE(DPT), \textcolor{blue}{-}: worse than USHADE(DPT), $\approx$: no significant difference)}
\label{tab:uadeT-D30}
\centering
\begin{tabular}{c|cc|cc|cc|c}
   & \multicolumn{2}{c|}{LSHADE} & \multicolumn{2}{c|}{SHADE} & \multicolumn{2}{c|}{USHADE(T)} & USHADE(DPT) \\\hline\hline
$F1 $ & 0.00E+00      & $\approx$     & 2.46E+02     & \textcolor{blue}{-}          & 0.00E+00       & $\approx$      & 0.00E+00 \\\hline
$F2 $ & 0.00E+00      & $\approx$     & 0.00E+00     & $\approx$     & 0.00E+00       & $\approx$      & 0.00E+00 \\\hline
$F3 $ & 0.00E+00      & $\approx$     & 0.00E+00     & $\approx$     & 0.00E+00       & $\approx$      & 0.00E+00 \\\hline
$F4 $ & 0.00E+00      & $\approx$     & 0.00E+00     & $\approx$     & 0.00E+00       & $\approx$      & 0.00E+00 \\\hline
$F5 $ & 2.00E+01      & \textcolor{blue}{-}          & 2.00E+01     & \textcolor{blue}{-}          & 2.04E+01       & \textcolor{blue}{-}           & 2.00E+01 \\\hline
$F6 $ & 3.19E-02      & $\approx$     & 1.08E+00     & \textcolor{blue}{-}          & 6.21E-02       & $\approx$      & 6.98E-02 \\\hline
$F7 $ & 0.00E+00      & $\approx$     & 2.90E-04     & $\approx$     & 0.00E+00       & $\approx$      & 0.00E+00 \\\hline
$F8 $ & 0.00E+00      & $\approx$     & 0.00E+00     & $\approx$     & 5.07E-01       & \textcolor{blue}{-}           & 0.00E+00 \\\hline
$F9 $ & 6.61E+00      & \textcolor{red}{+}          & 1.53E+01     & \textcolor{blue}{-}          & 1.47E+01       & \textcolor{blue}{-}           & 9.41E+00 \\\hline
$F10$ & 4.08E-04      & \textcolor{red}{+}          & 5.72E-03     & \textcolor{red}{+}          & 1.08E+01       & \textcolor{blue}{-}           & 6.53E-01 \\\hline
$F11$ & 1.07E+03      & \textcolor{red}{+}          & 1.33E+03     & \textcolor{blue}{-}          & 2.03E+03       & \textcolor{blue}{-}           & 1.12E+03 \\\hline
$F12$ & 1.01E-01      & \textcolor{blue}{-}          & 1.12E-01     & \textcolor{blue}{-}          & 1.89E-01       & \textcolor{blue}{-}           & 7.26E-02 \\\hline
$F13$ & 1.02E-01      & \textcolor{blue}{-}          & 1.86E-01     & \textcolor{blue}{-}          & 6.66E-02       & \textcolor{red}{+}           & 7.28E-02 \\\hline
$F14$ & 2.08E-01      & \textcolor{red}{+}          & 2.11E-01     & \textcolor{red}{+}          & 2.42E-01       & $\approx$      & 2.36E-01 \\\hline
$F15$ & 1.80E+00      & \textcolor{red}{+}          & 2.14E+00     & $\approx$     & 3.07E+00       & \textcolor{blue}{-}           & 2.23E+00 \\\hline
$F16$ & 8.13E+00      & \textcolor{blue}{-}          & 8.96E+00     & \textcolor{blue}{-}          & 8.73E+00       & \textcolor{blue}{-}           & 7.94E+00 \\\hline
$F17$ & 1.58E+01      & \textcolor{blue}{-}          & 2.00E+01     & \textcolor{blue}{-}          & 1.35E+01       & $\approx$      & 1.54E+01 \\\hline
$F18$ & 2.49E+00      & $\approx$     & 7.68E+00     & \textcolor{blue}{-}          & 1.97E+00       & $\approx$      & 4.10E+00 \\\hline
$F19$ & 2.24E+01      & \textcolor{blue}{-}          & 2.85E+01     & \textcolor{blue}{-}          & 2.51E+01       & \textcolor{blue}{-}           & 2.35E+01 \\\hline
$F20$ & 1.99E+01      & \textcolor{red}{+}          & 2.23E+01     & \textcolor{blue}{-}          & 2.07E+01       & $\approx$      & 2.14E+01 \\\hline
$F21$ & 6.53E+00      & \textcolor{blue}{-}          & 8.59E+00     & \textcolor{blue}{-}          & 6.60E+00       & $\approx$      & 6.09E+00 \\\hline
$F22$ & 1.84E+01      & \textcolor{red}{+}          & 1.87E+01     & \textcolor{red}{+}          & 2.08E+01       & \textcolor{blue}{-}           & 2.11E+01 \\\hline
$F23$ & 3.15E+02      & \textcolor{blue}{-}          & 3.15E+02     & \textcolor{blue}{-}          & 3.15E+02       & $\approx$      & 3.15E+02 \\\hline
$F24$ & 2.25E+02      & $\approx$     & 2.25E+02     & \textcolor{blue}{-}          & 2.25E+02       & \textcolor{blue}{-}           & 2.24E+02 \\\hline
$F25$ & 2.03E+02      & \textcolor{blue}{-}          & 2.06E+02     & \textcolor{blue}{-}          & 2.03E+02       & $\approx$      & 2.03E+02 \\\hline
$F26$ & 1.00E+02      & \textcolor{blue}{-}          & 1.02E+02     & \textcolor{blue}{-}          & 1.00E+02       & \textcolor{red}{+}           & 1.00E+02 \\\hline
$F27$ & 3.02E+02      & \textcolor{blue}{-}          & 3.38E+02     & \textcolor{blue}{-}          & 3.00E+02       & \textcolor{blue}{-}           & 3.01E+02 \\\hline
$F28$ & 8.19E+02      & \textcolor{red}{+}          & 8.36E+02     & $\approx$     & 8.33E+02       & $\approx$      & 8.36E+02 \\\hline
$F29$ & 3.13E+06      & $\approx$     & 4.07E+06     & \textcolor{blue}{-}          & 3.57E+06       & $\approx$      & 3.15E+06 \\\hline
$F30$ & 2.14E+02      & $\approx$     & 2.22E+02     & \textcolor{blue}{-}          & 2.18E+02       & $\approx$      & 2.18E+02 \\\hline
\end{tabular}
\end{table}

%% file: tab-alladd-D50.tex
\begin{table}[htbp]
  \caption{\small
Comparison of LSHADE, SHADE, USHADE(T) vs. USHADE(DPT)
(CEC 2014 benchmarks, 50 dimension, maximum evaluation budgets $=1,000,000$). Wilcoxon ranked
sum test(p = 0.05) results on $F1$ to $F30$ are shown.
(\textcolor{red}{+}: better than USHADE(DPT), \textcolor{blue}{-}: worse than USHADE(DPT), $\approx$: no significant difference)
}
\label{tab:uadeT-D50}
\centering
\begin{tabular}{c|cc|cc|cc|c}
   & \multicolumn{2}{c|}{LSHADE} & \multicolumn{2}{c|}{SHADE} & \multicolumn{2}{c|}{USHADE(T)} & USHADE(DPT) \\\hline\hline
$F1 $ & 1.59E+04      & \textcolor{blue}{-}          & 1.85E+04     & \textcolor{blue}{-}          & 1.34E+03       & $\approx$      & 9.91E+02 \\\hline
$F2 $ & 0.00E+00      & $\approx$     & 0.00E+00     & $\approx$     & 0.00E+00       & $\approx$      & 0.00E+00 \\\hline
$F3 $ & 0.00E+00      & $\approx$     & 5.85E-08     & \textcolor{blue}{-}          & 0.00E+00       & $\approx$      & 0.00E+00 \\\hline
$F4 $ & 1.02E+01      & \textcolor{blue}{-}          & 2.05E+00     & \textcolor{blue}{-}          & 1.92E+01       & $\approx$      & 2.31E+01 \\\hline
$F5 $ & 2.01E+01      & $\approx$     & 2.00E+01     & \textcolor{red}{+}          & 2.05E+01       & \textcolor{blue}{-}           & 2.01E+01 \\\hline
$F6 $ & 1.21E+00      & $\approx$     & 7.29E+00     & \textcolor{blue}{-}          & 1.70E+00       & $\approx$      & 1.44E+00 \\\hline
$F7 $ & 2.46E-03      & \textcolor{blue}{-}          & 5.99E-03     & \textcolor{blue}{-}          & 2.90E-04       & $\approx$      & 3.87E-04 \\\hline
$F8 $ & 0.00E+00      & $\approx$     & 0.00E+00     & $\approx$     & 4.70E+00       & \textcolor{blue}{-}           & 0.00E+00 \\\hline
$F9 $ & 1.41E+01      & \textcolor{red}{+}          & 3.66E+01     & \textcolor{blue}{-}          & 3.36E+01       & \textcolor{blue}{-}           & 2.14E+01 \\\hline
$F10$ & 1.96E-03      & \textcolor{red}{+}          & 7.59E-03     & \textcolor{red}{+}          & 8.90E+01       & \textcolor{blue}{-}           & 1.13E-02 \\\hline
$F11$ & 2.79E+03      & \textcolor{red}{+}          & 3.39E+03     & \textcolor{blue}{-}          & 5.17E+03       & \textcolor{blue}{-}           & 3.06E+03 \\\hline
$F12$ & 1.31E-01      & \textcolor{blue}{-}          & 1.10E-01     & \textcolor{red}{+}          & 3.78E-01       & \textcolor{blue}{-}           & 1.22E-01 \\\hline
$F13$ & 1.77E-01      & \textcolor{red}{+}          & 3.11E-01     & \textcolor{blue}{-}          & 1.75E-01       & \textcolor{red}{+}           & 2.06E-01 \\\hline
$F14$ & 2.88E-01      & $\approx$     & 2.78E-01     & \textcolor{red}{+}          & 2.94E-01       & $\approx$      & 2.89E-01 \\\hline
$F15$ & 4.19E+00      & \textcolor{red}{+}          & 5.55E+00     & \textcolor{blue}{-}          & 5.50E+00       & \textcolor{blue}{-}           & 4.57E+00 \\\hline
$F16$ & 1.62E+01      & \textcolor{red}{+}          & 1.71E+01     & \textcolor{blue}{-}          & 1.82E+01       & \textcolor{blue}{-}           & 1.66E+01 \\\hline
$F17$ & 2.03E+01      & $\approx$     & 2.04E+01     & \textcolor{blue}{-}          & 2.04E+01       & \textcolor{blue}{-}           & 2.03E+01 \\\hline
$F18$ & 6.92E+00      & \textcolor{blue}{-}          & 1.17E+01     & \textcolor{blue}{-}          & 8.13E+00       & $\approx$      & 6.23E+00 \\\hline
$F19$ & 4.77E+01      & \textcolor{blue}{-}          & 4.94E+01     & \textcolor{blue}{-}          & 3.61E+01       & $\approx$      & 3.30E+01 \\\hline
$F20$ & 2.34E+01      & \textcolor{red}{+}          & 2.81E+01     & \textcolor{blue}{-}          & 2.40E+01       & \textcolor{blue}{-}           & 2.38E+01 \\\hline
$F21$ & 1.23E+01      & \textcolor{red}{+}          & 1.78E+01     & \textcolor{blue}{-}          & 1.82E+01       & \textcolor{blue}{-}           & 1.28E+01 \\\hline
$F22$ & 2.17E+01      & \textcolor{red}{+}          & 2.26E+01     & $\approx$     & 2.27E+01       & $\approx$      & 2.25E+01 \\\hline
$F23$ & 3.44E+02      & \textcolor{blue}{-}          & 3.44E+02     & \textcolor{blue}{-}          & 3.44E+02       & $\approx$      & 3.44E+02 \\\hline
$F24$ & 2.75E+02      & \textcolor{blue}{-}          & 2.74E+02     & $\approx$     & 2.75E+02       & \textcolor{blue}{-}           & 2.74E+02 \\\hline
$F25$ & 2.06E+02      & $\approx$     & 2.22E+02     & \textcolor{blue}{-}          & 2.06E+02       & $\approx$      & 2.06E+02 \\\hline
$F26$ & 1.00E+02      & $\approx$     & 1.12E+02     & \textcolor{blue}{-}          & 1.00E+02       & \textcolor{red}{+}           & 1.02E+02 \\\hline
$F27$ & 3.31E+02      & \textcolor{red}{+}          & 4.69E+02     & \textcolor{blue}{-}          & 3.45E+02       & $\approx$      & 3.44E+02 \\\hline
$F28$ & 1.12E+03      & \textcolor{red}{+}          & 1.17E+03     & \textcolor{blue}{-}          & 1.15E+03       & \textcolor{blue}{-}           & 1.13E+03 \\\hline
$F29$ & 5.01E+04      & \textcolor{blue}{-}          & 5.06E+04     & \textcolor{blue}{-}          & 5.01E+04       & \textcolor{blue}{-}           & 5.01E+04 \\\hline
$F30$ & 2.34E+02      & $\approx$     & 2.38E+02     & \textcolor{blue}{-}          & 2.35E+02       & $\approx$      & 2.34E+02 \\\hline
\end{tabular}
\end{table}

%% file: tab-cec22comp.tex
\begin{table}[htbp]
  \caption{\small
Comparison of S-LSHADE-DP (4th place), NL-SHADE-RSP-MID (3rd place), NL-SHADE-LBC (2nd place) and EA4eig (1st place) vs USHADE(DPT).
(CEC 2022 benchmarks, 10 and 20 dimension, maximum evaluation budgets $=200,000$ for $D=10$ and $=1,000,000$ for $D=20$).
Wilcoxon ranked sum test($p = 0.05$) results on $F1$ to $F12$ are shown.
(\textcolor{blue}{-}: worse than USHADE(DPT), \textcolor{red}{+}: better than USHADE(DPT), $\approx$: no significant difference)
}
\label{tab:cec22}
\centering
\begin{tabular}{c|cc|cc|cc|cc|c}
$D=10$ & \multicolumn{2}{c|}{S-LSHADE-DP} & \multicolumn{2}{c|}{NL-SHADE-RSP-MID} & \multicolumn{2}{c|}{NL-SHADE-LBC} & \multicolumn{2}{c|}{EA4eig} & USHADE(DPT) \\\hline\hline
$F1 $ & 0E+00    & $\approx$ & 0E+00    & $\approx$ & 0E+00    & $\approx$ & 0E+00    & $\approx$ & 0E+00    \\\hline
$F2 $ & 0E+00    & \textcolor{red}{+}     & 0E+00    & \textcolor{red}{+}     & 1.33E-01 & \textcolor{red}{+}     & 1.46E+00 & \textcolor{red}{+}     & 6.15E+00 \\\hline
$F3 $ & 0E+00    & $\approx$ & 0E+00    & $\approx$ & 0E+00    & $\approx$ & 0E+00    & $\approx$ & 0E+00    \\\hline
$F4 $ & 4.72E+00 & \textcolor{blue}{-}     & 1E+01    & \textcolor{blue}{-}     & 1.3E+00  & \textcolor{red}{+}     & 1.26E+00 & \textcolor{red}{+}     & 3.18E+00 \\\hline
$F5 $ & 0E+00    & $\approx$ & 1.69E+00 & \textcolor{blue}{-}     & 0E+00    & $\approx$ & 0E+00    & $\approx$ & 0E+00    \\\hline
$F6 $ & 2.6E-01  & $\approx$ & 1.67E-01 & \textcolor{red}{+}     & 1.24E-01 & \textcolor{red}{+}     & 1.74E-02 & \textcolor{red}{+}     & 2.85E-01 \\\hline
$F7 $ & 0E+00    & \textcolor{red}{+}     & 0E+00    & \textcolor{red}{+}     & 0E+00    & \textcolor{red}{+}     & 0E+00    & \textcolor{red}{+}     & 1.25E-05 \\\hline
$F8 $ & 1.89E-01 & $\approx$ & 2.38E-01 & $\approx$ & 4.6E-02  & $\approx$ & 7.09E-02 & $\approx$ & 2.42E-01 \\\hline
$F9 $ & 1.91E+02 & \textcolor{blue}{-}     & 2.29E+02 & \textcolor{red}{+}     & 2.29E+02 & \textcolor{red}{+}     & 1.86E+02 & \textcolor{red}{+}     & 2.29E+02 \\\hline
$F10$ & 1.25E-02 & \textcolor{red}{+}     & 4.53E+00 & \textcolor{red}{+}     & 1E+02    & \textcolor{red}{+}     & 1E+02    & \textcolor{red}{+}     & 1E+02    \\\hline
$F11$ & 0E+00    & $\approx$ & 4.52E-10 & $\approx$ & 0E+00    & $\approx$ & 0E+00    & $\approx$ & 0E+00    \\\hline
$F12$ & 1.62E+02 & \textcolor{red}{+}     & 1.65E+02 & \textcolor{blue}{-}     & 1.65E+02 & \textcolor{blue}{-}     & 1.47E+02 & \textcolor{red}{+}     & 1.64E+02 \\\hline
\\
$D=20$ & \multicolumn{2}{c|}{S-LSHADE-DP} & \multicolumn{2}{c|}{NL-SHADE-RSP-MID} & \multicolumn{2}{c|}{NL-SHADE-LBC} & \multicolumn{2}{c|}{EA4eig} & USHADE(DPT) \\\hline\hline
$F1 $ & 0E+00    & $\approx$ & 0E+00    & $\approx$ & 0E+00    & $\approx$ & 0E+00    & $\approx$ & 0E+00    \\\hline
$F2 $ & 4.03E-01 & \textcolor{red}{+}     & 8.93E+00 & \textcolor{red}{+}     & 4.73E+01 & \textcolor{blue}{-}     & 1.06E+00 & \textcolor{red}{+}     & 4.87E+01 \\\hline
$F3 $ & 0E+00    & $\approx$ & 0E+00    & $\approx$ & 0E+00    & $\approx$ & 0E+00    & $\approx$ & 0E+00    \\\hline
$F4 $ & 1.34E+01 & \textcolor{blue}{-}     & 2.79E+01 & \textcolor{blue}{-}     & 4.45E+00 & \textcolor{red}{+}     & 8.69E+00 & $\approx$ & 8.46E+00 \\\hline
$F5 $ & 2.98E-03 & $\approx$ & 1.47E+02 & \textcolor{blue}{-}     & 0E+00    & $\approx$ & 0E+00    & $\approx$ & 0E+00    \\\hline
$F6 $ & 2.14E+00 & $\approx$ & 6.35E+00 & \textcolor{blue}{-}     & 6.36E-01 & \textcolor{red}{+}     & 1.49E-01 & \textcolor{red}{+}     & 1.22E+00 \\\hline
$F7 $ & 1.31E+01 & $\approx$ & 1.16E+01 & \textcolor{blue}{-}     & 2.58E+00 & \textcolor{red}{+}     & 3.5E+00  & $\approx$ & 8.64E+00 \\\hline
$F8 $ & 1.87E+01 & \textcolor{blue}{-}     & 2E+01    & $\approx$ & 1.65E+01 & $\approx$ & 1.66E+01 & $\approx$ & 1.56E+01 \\\hline
$F9 $ & 1.81E+02 & \textcolor{blue}{-}     & 1.81E+02 & \textcolor{red}{+}     & 1.81E+02 & \textcolor{red}{+}     & 1.65E+02 & \textcolor{red}{+}     & 1.81E+02 \\\hline
$F10$ & 0E+00    & \textcolor{red}{+}     & 2.08E-03 & \textcolor{red}{+}     & 1E+02    & \textcolor{red}{+}     & 1.08E+02 & \textcolor{red}{+}     & 1E+02    \\\hline
$F11$ & 3E+01    & \textcolor{red}{+}     & 1.5E+02  & \textcolor{red}{+}     & 3.03E+02 & $\approx$ & 3.23E+02 & \textcolor{blue}{-}     & 3E+02    \\\hline
$F12$ & 2.34E+02 & $\approx$ & 2.43E+02 & \textcolor{blue}{-}     & 2.39E+02 & \textcolor{blue}{-}     & 2E+02    & \textcolor{red}{+}     & 2.35E+02 \\\hline
\end{tabular}
\end{table}

%% file: sec5.tex
\subsection{Simulating adaptive population sizes} \label{simulating-popsize}
The adaptive control of the tournament parameter $T$ in USHADE (Section \ref{sec:adaptive-selection-policy-parameter} is effectively similar to adaptively controlling population size in standard generational replacement-based DEs.
 Increasing the tournament size in the T and DPT policies (i.e., decreasing the parameter $T^{i,t}$) results in the selection of a smaller number of better-fitness individuals for offspring generation. Although the probability of selection varies among individuals, using only better-fitness individuals for mutation closely resembles reducing the population size in standard DE by eliminating worse-fitness individuals. Conversely, decreasing the tournament size (increasing $T^{i,t}$) has an effect similar to re-including previously excluded/discarded worse-fitness individuals for mutation, and is analogous to enlarging the population.

\subsubsection{Behavior of adaptively controlled $T$ parameter}
 
\begin{figure}[htbp]
\centering
\includegraphics[width=\textwidth]{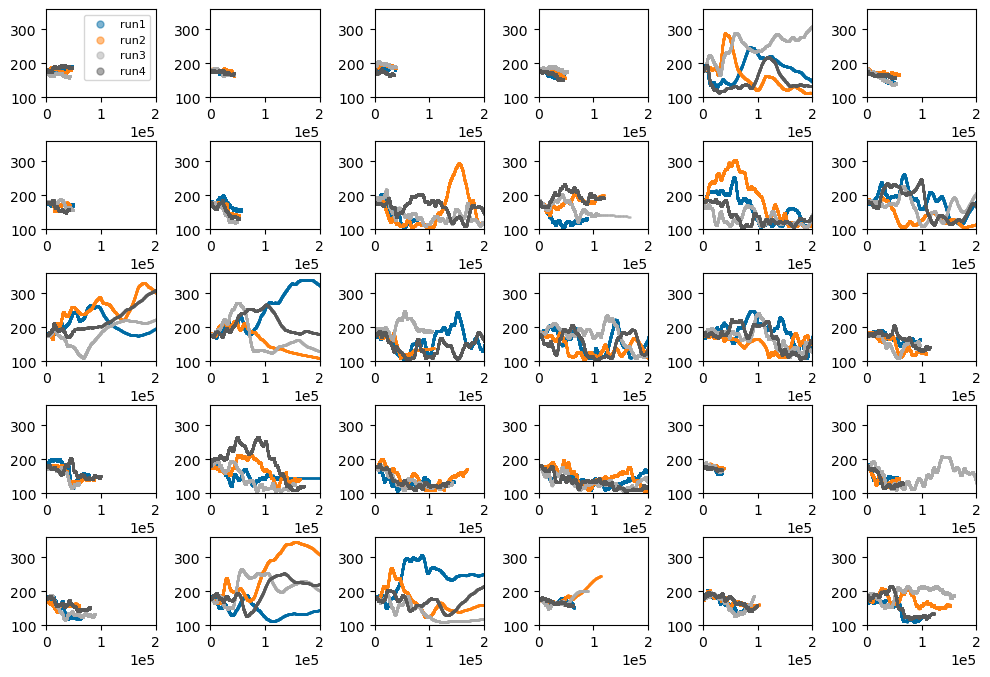}
\caption{\small
  Values of adaptive parameter $T$ in USHADE(DPT) for the CEC 2014 benchmarks with $D=10$. The curves represent moving averages of $T$ with a window width of $18\times D$, 4 independent trials per problem (each trial is a colored line). The vertical axis corresponds to $T$, which is initialized to $18\times D$ and has a minimum value of 100. The horizontal axis denotes the number of fitness evaluations.
The subfigures (1 subfigure/problem) are arranged in order 
 (Row 1:$F1-F6$, Row 2:$F7-F12$,..., Row 6: $F25-F30$).
}
\label{fig:parameter_D10}
\end{figure}

\begin{figure}[htbp]
\centering
$D=30$\\
\includegraphics[width=0.9\textwidth]{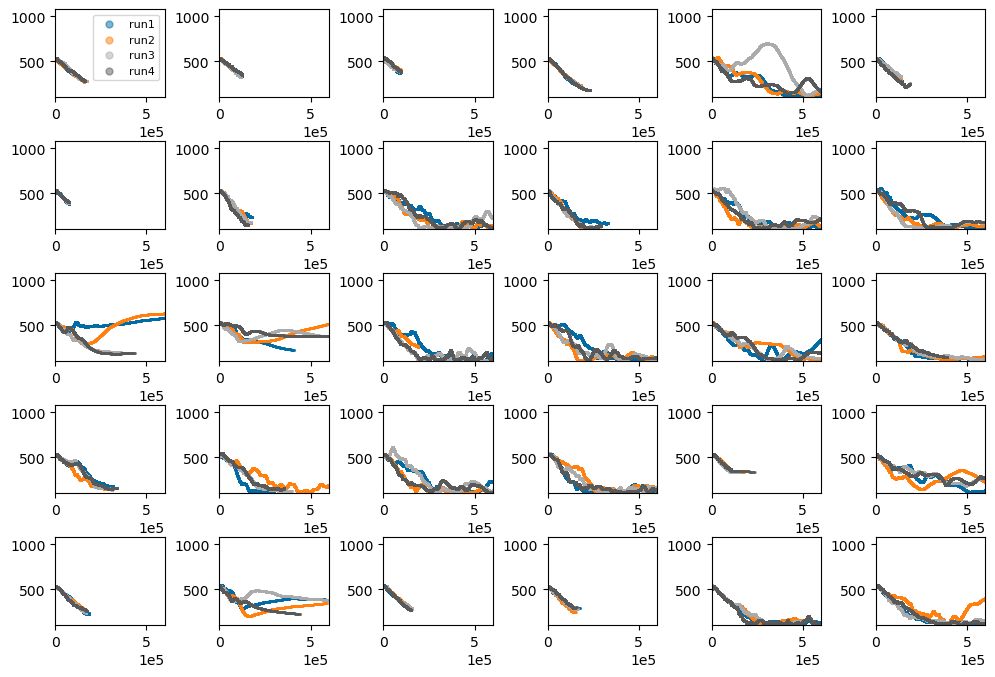}\\
$D=50$\\
\includegraphics[width=0.9\textwidth]{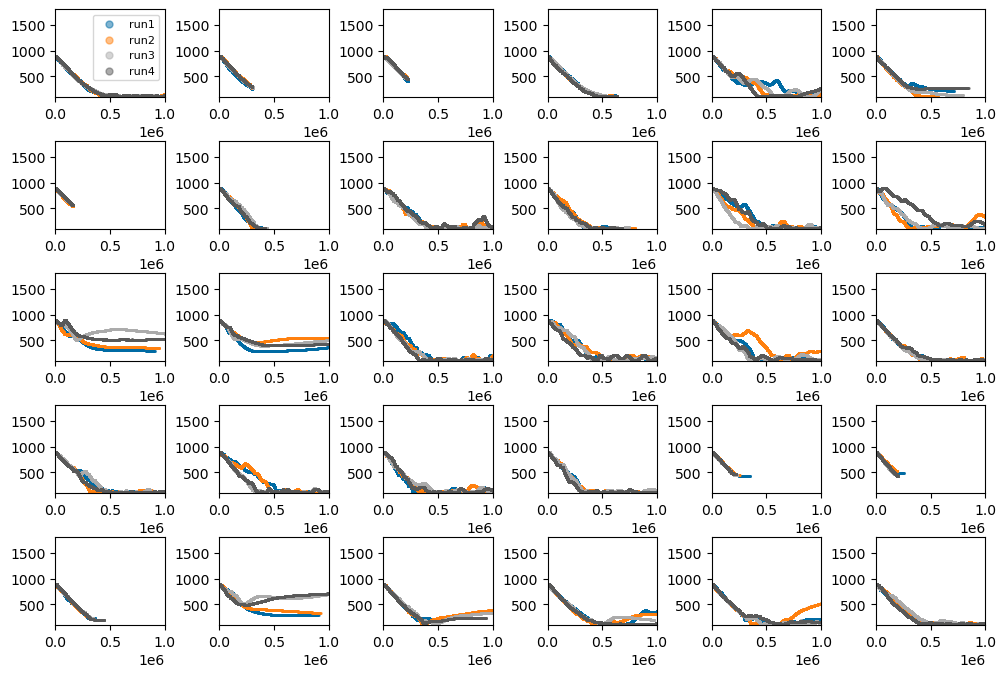}
\caption{\small
  Values of adaptive parameter $T$ in USHADE(DPT) for the CEC 2014 benchmarks with $D=30$ and $D=50$. The curves represent moving averages of $T$ with a window width of $18\times D$, 4 independent trials per problem (each trial is a colored line). The vertical axis corresponds to $T$, which is initialized to $18\times D$ and has a minimum value of 100. The horizontal axis denotes the number of fitness evaluations.
The subfigures (1 subfigure/problem) are arranged in order 
 (Row 1:$F1-F6$, Row 2:$F7-F12$,..., Row 6: $F25-F30$).
}
\label{fig:parameter_D30}
\end{figure}

To understand how the adaptive control of $T$ in USHADE(DPT) behaves, we plot smoothed trajectories of $T^{i,t}$ for all of the CEC2014 benchmarks (4 runs/problem), shown in Figures \ref{fig:parameter_D10} and \ref{fig:parameter_D30}.
Each colored line corresponds to 1 run. Across all problems, $T^{i,t}$ tends to decrease in the early search phase, with many runs reaching the minimum threshold, after which $T^{i,t}$ generally increases. Exceptions include benchmarks $F1$, $F2$, and $F3$, where the optimal solution is found early, and no subsequent increase in $T^{i,t}$  occurs.

The behavior of parameter $T$ varies considerably across different problems.
For instance, in the case of $D=10$ (Figure \ref{fig:parameter_D10}, problems such as $F21$ and $F22$ exhibit relatively stable values of $T$, whereas problems like $F5, F14, F26$, and $F27$ showed large fluctuations. In the case of $D=30$ (Figure \ref{fig:parameter_D30}, top), $T$ rarely exceeds its initial value, with the only exceptions occurring in problems $F5, F13$, and $F14$. Similarly, for $D=50$ (Figure \ref{fig:parameter_D30}, bottom), problems $F13$ and $F14$ are distinctive in that the value of $T$ levels off around 500, in contrast to other problems. Thus, the behavior of $T$ varies significantly depending on the problem, suggesting that the $T$  is being adaptively controlled in response to problem characteristics.

\subsubsection{Robustness with respect to maximum evaluation budgets}
\label{sec:robustness-evaluation-budgets}
If adaptive control of $T$ is effectively behaving like adaptive population control, then a possible, significant benefit is that unlike standard, deterministic control of the population size, such as the widely used linear population size reduction method (LPSR), whose performance depends on tuning a predefined reduction schedule (reducing the population size from $|P^{1}|$ to $|P|_\text{min}$ linearly) to the maximum evaluation budgets $L^\text{evaluation}_\text{max}$ (equation~\ref{eq:LPSR}) which is known {\it a priori}, the performance of USHADE should be somewhat robust with respect to the maximum evaluation budgets.

\begin{figure}[htbp]
\begin{center}
\includegraphics[width=.92\textwidth]{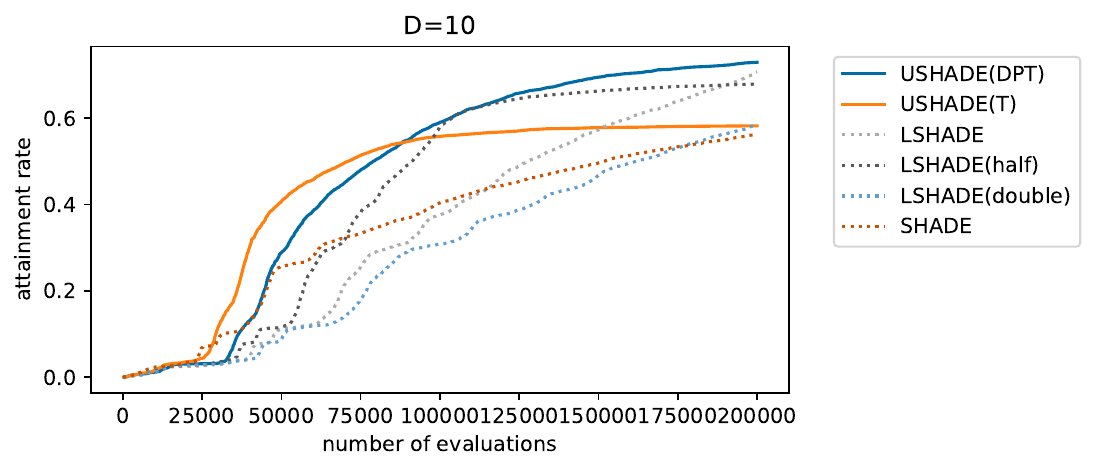}
\includegraphics[width=.92\textwidth]{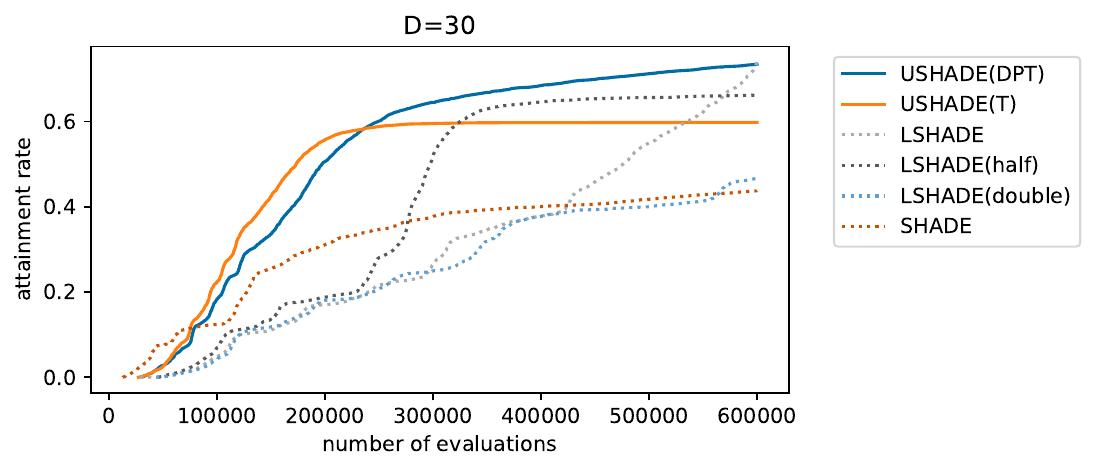}
\includegraphics[width=.92\textwidth]{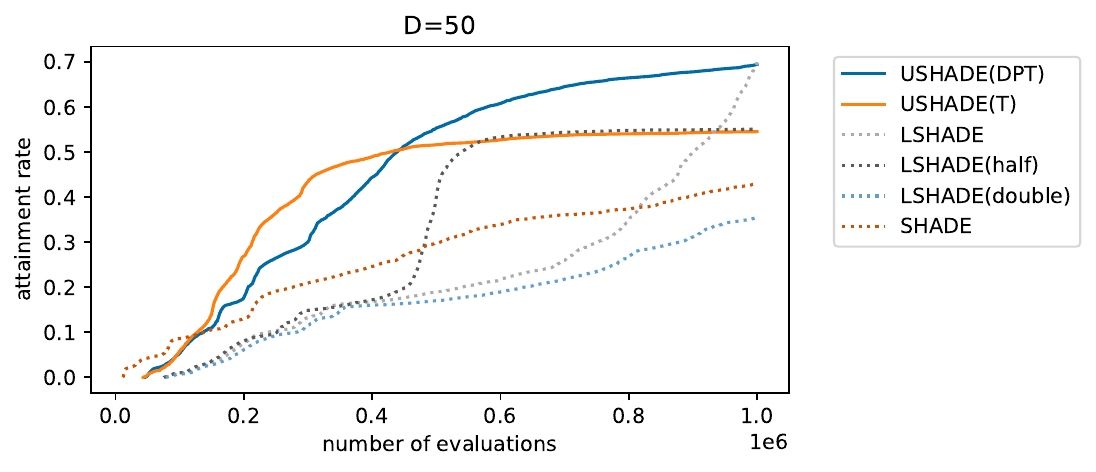}
\end{center}
\caption{\small
Empirical cumulative distribution functions (ECDFs) for six algorithms—USHADE(T), USHADE(DPT), SHADE, and three variants of LSHADE (standard, half schedule, and double schedule)—on the CEC 2014 benchmarks with $D=10,30,50$. In each plot, the horizontal axis indicates the number of fitness evaluations, while the vertical axis shows the proportion of runs (out of 51) in which the $\text{fitness}_\text{best-so-far}$ reached a predefined target. These targets were determined per problem based on the median and the first and third quartiles of $\FinalBestSoFar$, calculated across all 51 runs and all six algorithms. The ECDF shown represents the average across all 30 benchmark problems.}
\label{fig:lshades}
\end{figure}

Figure \ref{fig:lshades} presents the ECDFs of USHADE(T), USHADE(DPT), SHADE, and three variants of LSHADE on the CEC 2014 benchmark suite for $D=10,30,50$. The linear population size reduction (LPSR) strategy employed by LSHADE requires a predefined reduction schedule.
\begin{itemize}
\item The standard version of LSHADE reaches its minimum population size $|P|_\text{min}=4$ after $maxevals = L^\text{evaluation}_\text{max}$ evaluations.
\item LSHADE (half) is a variant scheduled to reach $|P|_\text{min}$ at $maxevals = L^\text{evaluation}_\text{max}/2$ evaluations.
\item LSHADE (double) uses a schedule which reaches $|P|_\text{min}$ after $maxevals = 2 \times L^\text{evaluation}_\text{max}$  evaluations.
\end{itemize}

As shown in the figure, %
LSHADE(half) shows a rapid increase in ECDF as the scheduled $maxevals$ approach and a very slow pace of increase again after $maxevals$ is exceeded.
Thus, LSHADE (standard) has an ECDF that increases as the number of fitness evaluations approaches $L^\text{evaluation}_\text{max}$, and if the search continues after that, performance is expected to stagnate.
LSHADE (double) with $maxevals$ larger than $L^\text{evaluation}_\text{max}$ is not performing well due to improper scheduling.
LPSR of LSHADE makes it difficult to adjust the evaluation budgets dynamically.
In contrast, USHADE implicitly mimics population size control through adaptive mechanisms, eliminating the need for such scheduling.

Figure \ref{fig:lshades} shows that USHADE outperforms LSHADE, LSHADE (half), and LSHADE (double). Thus, adaptive control of $T$ in USHADE, enabled by the flexibility afforded by the UDE framework which keeps all created individuals, is a significant advantage over the standard generational replacement based DE which discards valuable information (failed individuals, as well as individuals replaced during generational replacement).

\clearpage
\subsection{Are failed individuals useful or harmful?}  \label{sec:ex2}
Standard DE algorithms only keep successful individuals (individuals with better fitness than their parent) in the population, and discard failed individuals. In contrast, USHADE keeps all offspring in the population, including failed ones. This section empirically investigates whether failed individuals contribute to search progress in USHADE, and whether excluding failed individuals affects convergence speed to better-fitness solutions.

\subsubsection{Frequency of Usage of Failed Individuals} \label{sec:usage} %
We analyzed whether the best-so-far offspring at any given point during a single run of the CEC 2014 benchmarks was derived from a failed or successful parent.
As individuals were created, we marked them as ``successful'' or ``failed'' depending on whether their fitness was better than their parent, and 
each time a new best-so-far individual was evaluated (when $\text{fitness}_\text{best-so-far}$ is updated), we recorded whether its parent was a successful individual or failed individual.

\begin{figure}[htbp]
\begin{center}
\includegraphics[width=1\textwidth]{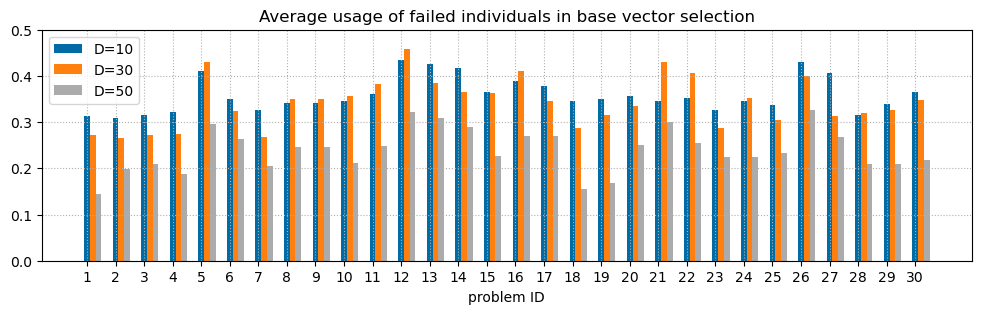}
\end{center}
\caption{\small
Fraction of best-so-far individuals found during search whose parent was a failed individual (30 problems from CEC 2014 benchmarks, 51 runs/problem, $D=10,30,50$).}
\label{fig:usage}
\end{figure}

Figure \ref{fig:usage} shows the fraction of best-so-far individuals that had a failed parent, for each of the 30 benchmark problems ($D=10, 30, 50$ dimensions) across 51 runs for each problem. The fraction of best-so-far individuals with failed parents was approximately 30\% for $D=10, 30$, and around 20\% for $D=50$.
Since a significant proportion of the best individuals were generated from failed parents, this indicates that in USHADE, failed individuals are not useless -- they contribute to search progress by being the parents of new best-so-far individuals.

Figure \ref{fig:usage} also shows that although the extent to which failed individuals are utilized in UDE varies depending on the problem, even the lowest utilization (for $D=50$ on the $F1$ problem) is around 15\%.

\subsubsection{Comparison of search progress with and without failed individuals} \label{sec:no-use}
Next, we evaluate whether failed individuals contribute positively or negatively to the search efficiency of USHADE.
We compare USHADE(DPT) to USHADE/DF (USHADE which discards all failed individual, defined in Section \ref{sec:ude-without-failed-individuals}).

\begin{figure}[htbp]
\begin{center}
\includegraphics[width=.9\textwidth]{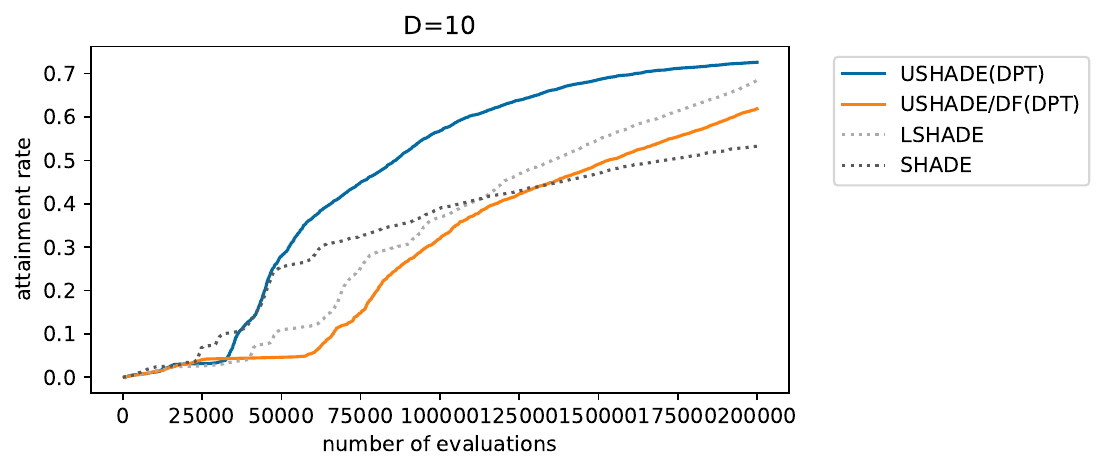}
\includegraphics[width=.9\textwidth]{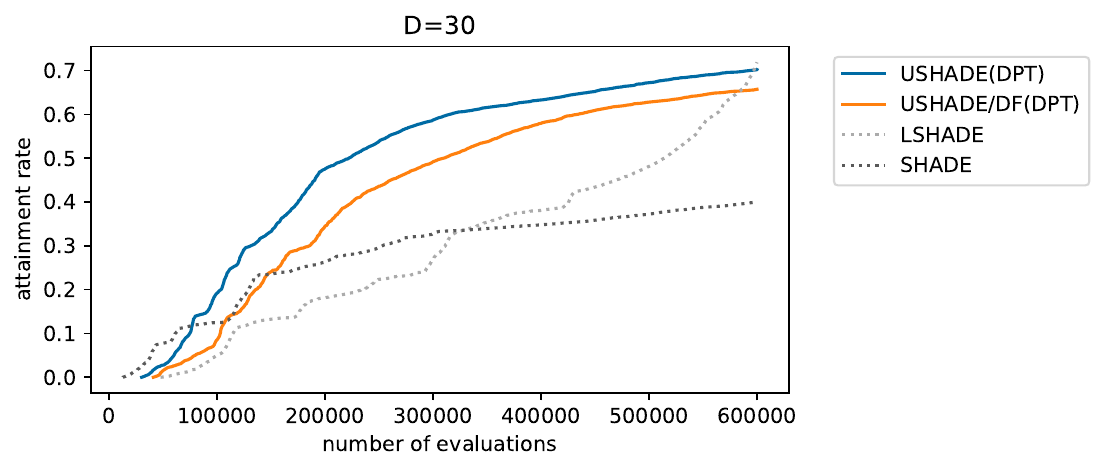}
\includegraphics[width=.9\textwidth]{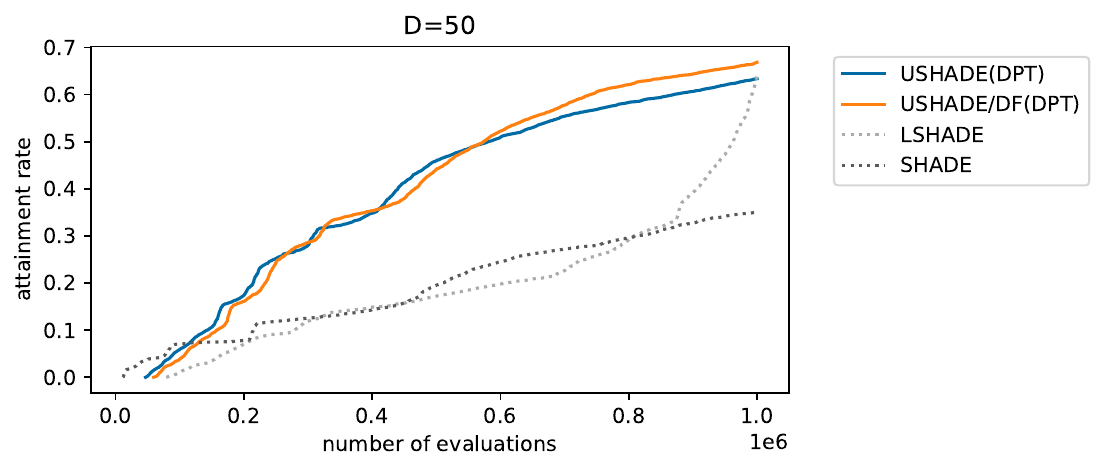}
\end{center}
\caption{\small
Empirical cumulative distribution functions (ECDFs) for four algorithms-USHADE(DPT), USHADE(DPT) without failed individual, LSHADE and SHADE—on the CEC 2014 benchmarks with $D=10,30,50$. In each plot, the horizontal axis indicates the number of fitness evaluations, while the vertical axis shows the proportion of runs (out of 51) in which the $\text{fitness}_\text{best-so-far}$ reached a predefined target. These targets were determined per problem based on the median and the first and third quartiles of $\FinalBestSoFar$, calculated across all 51 runs and all six algorithms. The ECDF shown represents the average across all 30 benchmark problems.}
\label{fig:wo-failed}
\end{figure}

Figure \ref{fig:wo-failed} compares the ECDFs for USHADE(DPT) and USHADE/DF(DPT). For $D=10$ and $D=30$, USHADE(DPT) consistently outperformed USHADE/DF(DPT). However, for $D=50$, the two variants performed comparably up to 50 million evaluations, after which USHADE/DF(DPT)  achieved slightly better performance near the end of the run.

\input tab_wo-failed

Tables~\ref{tab:wo_failed1} and \ref{tab:wo_failed2}
 comparing the $\text{fitness}_\text{best-so-far}$ of the two methods at the mid-point ($1e4\times D$) and endpoint ($2e4\times D$) of the search (Wilcoxon's rank-sum comparison, $p=0.05$)
Table~\ref{tab:wo_failed1} shows that at the midpoint of search ($1e4\times D$ evaluations), USHADE/DF(DPT) performed significantly worse than USHADE(DPT) on 22 problems for $D=10$, 17 problems for $D=30$, and 16 problems for $D=50$, indicating that including failed individuals helps find better solutions with fewer evaluations. However, for problems such as $F6$ and those from $F24$ onwards, USHADE/DF(DPT) achieved better fitness than USHADE(DPT). Since problems $F22$ and onward are composite functions that combine problems with different characteristics, this suggests that including failed individuals may hinder performance on more complex problems.

Table~\ref{tab:wo_failed2} shows that at the end of search  ($2e4\times D$ evaluations), the number of problems where USHADE/DF(DPT) outperformed USHADE(DPT)  increased to 5 for $D=10$, 9 for $D=30$, and 12 for $D=50$. These results suggest that although using failed individuals accelerates early-stage convergence, excluding them to focus more on the promising regions of the search can improve final solution quality given a sufficiently large evaluation budget.
However, this does not necessarily mean that discarding failed individuals is necessary -- it may be possible to improve the selection policy to increase the bias for selecting successful individuals as search progresses. This is a direction for future work.

%% file: tab_wo-failed.tex
\begin{table}[htbp]
  \caption{\small
Comparison of USHADE(DPT) vs. USHADE/DF(DPT) (USHADE without failed individuals) after $=10,000 \times D$ evaluations
(CEC 2014 benchmarks, 10,30,50 dimension). Wilcoxon ranked
sum test($p = 0.05$) results on $F1$ to $F30$ are shown.
(\textcolor{red}{+}: better than USHADE(DPT), \textcolor{blue}{-}: worse than USHADE(DPT), $\approx$: no significant difference)}
\label{tab:wo_failed1}
\centering
\begin{tabular}{c|ccc|ccc|ccc}
   & \multicolumn{3}{c|}{D=10}                       & \multicolumn{3}{c|}{D=30}                       & \multicolumn{3}{c}{D=50}                       \\ \hline
   & \multicolumn{2}{l}{w/o failed ind.} & USHADE(DPT) & \multicolumn{2}{l}{w/o failed ind.} & USHADE(DPT) & \multicolumn{2}{l}{w/o failed ind.} & USHADE(DPT) \\ \hline\hline

$F1 $ & 0                 & $\approx$          & 0        & 0                 & $\approx$          & 0        & 602.9064          & \textcolor{red}{+}               & 6254.75  \\ \hline
$F2 $ & 0                 & $\approx$          & 0        & 0                 & $\approx$          & 0        & 0                 & $\approx$          & 0        \\ \hline
$F3 $ & 0                 & $\approx$          & 0        & 0                 & $\approx$          & 0        & 0                 & $\approx$          & 0        \\ \hline
$F4 $ & 27.36365          & $\approx$          & 31.37044 & 0                 & $\approx$          & 0        & 26.35753          & $\approx$          & 24.60044 \\ \hline
$F5 $ & 18.63541          & \textcolor{blue}{-}               & 18.44447 & 20.26397          & \textcolor{blue}{-}               & 20.1613  & 20.41052          & \textcolor{blue}{-}               & 20.35465 \\ \hline
$F6 $ & 3.83E\textcolor{blue}{-}05          & \textcolor{blue}{-}               & 0        & 0.019293          & \textcolor{red}{+}               & 0.069785 & 0.353142          & \textcolor{red}{+}               & 1.439085 \\ \hline
$F7 $ & 0.05226           & \textcolor{blue}{-}               & 0.000822 & 0                 & $\approx$          & 0        & 0.000145          & $\approx$          & 0.000387 \\ \hline
$F8 $ & 5.29E\textcolor{blue}{-}07          & \textcolor{blue}{-}               & 0        & 0                 & $\approx$          & 0        & 0.000167          & \textcolor{blue}{-}               & 0        \\ \hline
$F9 $ & 3.508477          & \textcolor{blue}{-}               & 2.32013  & 13.91448          & \textcolor{blue}{-}               & 9.854631 & 29.99731          & \textcolor{blue}{-}               & 26.15617 \\ \hline
$F10$ & 0.719729          & \textcolor{blue}{-}               & 0.080744 & 2.437133          & \textcolor{blue}{-}               & 0.678128 & 3.517863          & \textcolor{blue}{-}               & 0.483164 \\ \hline
$F11$ & 81.96778          & \textcolor{blue}{-}               & 33.10976 & 1485.592          & \textcolor{blue}{-}               & 1193.491 & 4005.837          & \textcolor{blue}{-}               & 3608.658 \\ \hline
$F12$ & 0.29655           & \textcolor{blue}{-}               & 0.142306 & 0.251878          & \textcolor{blue}{-}               & 0.144858 & 0.319648          & \textcolor{blue}{-}               & 0.247534 \\ \hline
$F13$ & 0.083985          & \textcolor{blue}{-}               & 0.047563 & 0.117748          & \textcolor{blue}{-}               & 0.083674 & 0.210042          & $\approx$          & 0.216014 \\ \hline
$F14$ & 0.126246          & \textcolor{red}{+}               & 0.148488 & 0.237112          & $\approx$          & 0.237477 & 0.288249          & $\approx$          & 0.292059 \\ \hline
$F15$ & 0.68132           & \textcolor{blue}{-}               & 0.486435 & 2.961803          & \textcolor{blue}{-}               & 2.522984 & 6.745137          & \textcolor{blue}{-}               & 6.01941  \\ \hline
$F16$ & 1.513953          & \textcolor{blue}{-}               & 0.873816 & 8.842077          & \textcolor{blue}{-}               & 8.237343 & 17.52571          & \textcolor{blue}{-}               & 17.2602  \\ \hline
$F17$ & 0.084094          & \textcolor{blue}{-}               & 0.031245 & 11.51968          & \textcolor{blue}{-}               & 15.44986 & 20.64772          & \textcolor{blue}{-}               & 20.46071 \\ \hline
$F18$ & 0.097587          & \textcolor{blue}{-}               & 0.021522 & 5.805607          & \textcolor{blue}{-}               & 4.266267 & 8.489748          & \textcolor{blue}{-}               & 7.255026 \\ \hline
$F19$ & 0.044165          & \textcolor{blue}{-}               & 0.032618 & 24.65249          & \textcolor{blue}{-}               & 23.48928 & 45.0209           & \textcolor{blue}{-}               & 33.37428 \\ \hline
$F20$ & 0.234637          & \textcolor{blue}{-}               & 0.036678 & 21.72812          & \textcolor{blue}{-}               & 21.45243 & 24.39536          & \textcolor{blue}{-}               & 24.0944  \\ \hline
$F21$ & 3.066567          & \textcolor{blue}{-}               & 0.278178 & 9.071104          & \textcolor{blue}{-}               & 6.745998 & 17.41177          & \textcolor{blue}{-}               & 14.52838 \\ \hline
$F22$ & 0.28854           & \textcolor{blue}{-}               & 0.101654 & 18.27311          & \textcolor{blue}{-}               & 21.21274 & 22.90995          & $\approx$          & 22.98167 \\ \hline
$F23$ & 329.4575          & $\approx$          & 329.4575 & 315.2441          & \textcolor{blue}{-}               & 315.2441 & 344.0045          & $\approx$          & 344.0045 \\ \hline
$F24$ & 107.8189          & \textcolor{blue}{-}               & 107.2475 & 223.6052          & \textcolor{red}{+}               & 224.0858 & 273.9044          & $\approx$          & 273.6658 \\ \hline
$F25$ & 135.2816          & $\approx$          & 150.5073 & 202.5948          & \textcolor{red}{+}               & 202.6645 & 205.422           & \textcolor{red}{+}               & 205.6028 \\ \hline
$F26$ & 100.0897          & \textcolor{blue}{-}               & 100.0489 & 100.1201          & \textcolor{blue}{-}               & 100.0838 & 100.1811          & \textcolor{blue}{-}               & 102.124  \\ \hline
$F27$ & 19.1663           & \textcolor{blue}{-}               & 77.24684 & 300               & \textcolor{red}{+}               & 300.6497 & 326.4559          & \textcolor{red}{+}               & 343.8198 \\ \hline
$F28$ & 378.8699          & $\approx$          & 382.8362 & 829.358           & $\approx$          & 835.9293 & 1121.277          & \textcolor{red}{+}               & 1133.456 \\ \hline
$F29$ & 289.3911          & \textcolor{blue}{-}               & 375.2942 & 3534271           & \textcolor{blue}{-}               & 3153176  & 50119.63          & \textcolor{red}{+}               & 50123.85 \\ \hline
$F30$ & 160.4463          & \textcolor{blue}{-}               & 158.7343 & 215.1315          & $\approx$          & 217.982  & 234.038           & \textcolor{red}{+}               & 234.4386 \\\hline
\end{tabular}
\end{table}

\begin{table}[htbp]
  \caption{\small
Comparison of USHADE(DPT) vs. USHADE/DF(DPT) (USHADE without failed individuals) after $=20,000 \times D$ evaluations 
(CEC 2014 benchmarks, 10,30,50 dimension). Wilcoxon ranked
sum test($p = 0.05$) results on $F1$ to $F30$ are shown.
(\textcolor{red}{+}: better than USHADE(DPT), \textcolor{blue}{-}: worse than USHADE(DPT), $\approx$: no significant difference)}
\label{tab:wo_failed2}
\centering
\begin{tabular}{c|ccc|ccc|ccc}
   & \multicolumn{3}{c|}{D=10}                       & \multicolumn{3}{c|}{D=30}                       & \multicolumn{3}{c}{D=50}                       \\ \hline
   & \multicolumn{2}{l}{w/o failed ind.} & USHADE(DPT) & \multicolumn{2}{l}{w/o failed ind.} & USHADE(DPT) & \multicolumn{2}{l}{w/o failed ind.} & USHADE(DPT) \\ \hline\hline
1  & 0                 & $\approx$         & 0        & 0                 & $\approx$         & 0        & 66.1585           & \textcolor{red}{+}               & 991.2481 \\\hline
2  & 0                 & $\approx$         & 0        & 0                 & $\approx$         & 0        & 0                 & $\approx$         & 0        \\\hline
3  & 0                 & $\approx$         & 0        & 0                 & $\approx$         & 0        & 0                 & $\approx$         & 0        \\\hline
4  & 27.36365          & $\approx$         & 31.37044 & 0                 & $\approx$         & 0        & 25.01244          & $\approx$         & 23.08885 \\\hline
5  & 16.03121          & \textcolor{blue}{-}               & 16.02419 & 20.07797          & \textcolor{blue}{-}               & 20.02891 & 20.15423          & \textcolor{blue}{-}               & 20.12348 \\\hline
6  & 0.00E+00          & $\approx$         & 0        & 0.019293          & \textcolor{red}{+}               & 0.069785 & 0.353142          & \textcolor{red}{+}               & 1.439085 \\\hline
7  & 0.010306          & \textcolor{blue}{-}               & 0.000822 & 0                 & $\approx$         & 0        & 0.000145          & $\approx$         & 0.000387 \\\hline
8  & 0.00E+00          & $\approx$         & 0        & 0                 & $\approx$         & 0        & 0                 & $\approx$         & 0        \\\hline
9  & 2.501688          & $\approx$         & 2.282857 & 12.11628          & \textcolor{blue}{-}               & 9.407696 & 21.17888          & $\approx$         & 21.40621 \\\hline
10 & 0.058781          & $\approx$         & 0.079599 & 0.270126          & \textcolor{red}{+}               & 0.652983 & 0.002939          & \textcolor{red}{+}               & 0.011267 \\\hline
11 & 25.19619          & $\approx$         & 26.32415 & 1190.118          & $\approx$         & 1117.354 & 3120.054          & $\approx$         & 3055.083 \\\hline
12 & 0.107757          & \textcolor{blue}{-}               & 0.055584 & 0.102849          & \textcolor{blue}{-}               & 0.072594 & 0.142845          & \textcolor{blue}{-}               & 0.121509 \\\hline
13 & 0.049081          & \textcolor{blue}{-}               & 0.031336 & 0.104262          & \textcolor{blue}{-}               & 0.072806 & 0.198241          & \textcolor{red}{+}               & 0.206416 \\\hline
14 & 0.085738          & \textcolor{red}{+}               & 0.135742 & 0.226465          & \textcolor{red}{+}               & 0.235536 & 0.280785          & $\approx$         & 0.289214 \\\hline
15 & 0.4278            & $\approx$         & 0.451732 & 2.350371          & \textcolor{blue}{-}               & 2.226454 & 4.724278          & \textcolor{blue}{-}               & 4.569517 \\\hline
16 & 0.790968          & \textcolor{blue}{-}               & 0.667866 & 8.234898          & \textcolor{blue}{-}               & 7.940735 & 16.73666          & $\approx$         & 16.60553 \\\hline
17 & 0.036702          & \textcolor{blue}{-}               & 0.01572  & 11.10502          & $\approx$         & 15.38426 & 20.21586          & \textcolor{red}{+}               & 20.3113  \\\hline
18 & 0.015241          & \textcolor{red}{+}               & 0.021146 & 2.4102            & \textcolor{red}{+}               & 4.097872 & 6.80835           & $\approx$         & 6.225957 \\\hline
19 & 0.007803          & \textcolor{red}{+}               & 0.032618 & 23.60511          & $\approx$         & 23.45095 & 35.80443          & $\approx$         & 33.02532 \\\hline
20 & 0.033261          & \textcolor{blue}{-}               & 0.014619 & 20.354            & \textcolor{red}{+}               & 21.35249 & 23.47516          & \textcolor{red}{+}               & 23.7652  \\\hline
21 & 0.393259          & \textcolor{blue}{-}               & 0.008716 & 6.948644          & \textcolor{blue}{-}               & 6.091905 & 14.35929          & \textcolor{blue}{-}               & 12.81746 \\\hline
22 & 0.019624          & \textcolor{red}{+}               & 0.058067 & 17.3235           & \textcolor{red}{+}               & 21.05106 & 21.9324           & \textcolor{red}{+}               & 22.52048 \\\hline
23 & 329.4575          & $\approx$         & 329.4575 & 315.2441          & \textcolor{blue}{-}               & 315.2441 & 344.0045          & $\approx$         & 344.0045 \\\hline
24 & 107.1508          & $\approx$         & 107.2006 & 223.5992          & \textcolor{red}{+}               & 224.0849 & 273.9044          & $\approx$         & 273.6658 \\\hline
25 & 134.8596          & \textcolor{red}{+}               & 150.4212 & 202.5948          & \textcolor{red}{+}               & 202.6645 & 205.422           & \textcolor{red}{+}               & 205.6028 \\\hline
26 & 100.0561          & \textcolor{blue}{-}               & 100.0324 & 100.1065          & \textcolor{blue}{-}               & 100.0724 & 100.1697          & \textcolor{blue}{-}               & 102.1158 \\\hline
27 & 18.76087          & \textcolor{blue}{-}               & 77.145   & 300               & \textcolor{red}{+}               & 300.6497 & 326.4559          & \textcolor{red}{+}               & 343.8198 \\\hline
28 & 378.8699          & $\approx$         & 382.8362 & 829.358           & $\approx$         & 835.9293 & 1121.276          & \textcolor{red}{+}               & 1133.456 \\\hline
29 & 289.1087          & \textcolor{blue}{-}               & 375.2918 & 3534270           & $\approx$         & 3153176  & 50112.87          & \textcolor{red}{+}               & 50120.4  \\\hline
30 & 158.8163          & \textcolor{blue}{-}               & 158.7011 & 214.5777          & $\approx$         & 217.9746 & 233.8956          & \textcolor{red}{+}               & 234.4347 \\\hline
\end{tabular}
\end{table}